# Table2Vec: Automated Universal Representation Learning to Encode All-round Data DNA for Benchmarkable and Explainable Enterprise Data Science

**Longbing Cao*** **and Chengzhang Zhu**

Data Science Lab, University of Technology Sydney, Australia
*Longbing.Cao@uts.edu.au

## ABSTRACT

Enterprise data typically involves multiple heterogeneous data sources and external data that respectively record business activities, transactions, customer demographics, status, behaviors, interactions and communications with the enterprise, and the consumption and feedback of its products, services, production, marketing, operations, and management, etc. They involve *enterprise DNA* associated with domain-oriented transactions and master data, informational and operational metadata, and relevant external data. A critical challenge in enterprise data science is to enable an effective 'whole-of-enterprise' data understanding and data-driven discovery and decision-making on all-round enterprise DNA. Accordingly, here we introduce a neural encoder Table2Vec for *automated universal representation learning* of entities such as customers from all-round enterprise DNA with automated data characteristics analysis and data quality augmentation. The learned universal representations serve as representative and benchmarkable enterprise data genomes (similar to biological genomes and DNA in organisms) and can be used for enterprise-wide and domain-specific learning tasks. Table2Vec integrates automated universal representation learning on low-quality enterprise data and downstream learning tasks. Such automated universal enterprise representation and learning cannot be addressed by existing enterprise data warehouses (EDWs), business intelligence and corporate analytics systems, where 'enterprise big tables' are constructed with reporting and analytics conducted by specific analysts on respective domain subjects and goals. It addresses critical limitations and gaps of existing representation learning, enterprise analytics and cloud analytics, which are analytical subject, task and data-specific, creating *analytical silos* in an enterprise. We illustrate Table2Vec in characterizing all-round customer data DNA in an enterprise on complex heterogeneous multi-relational big tables to build universal customer vector representations. The learned universal representation of each customer is all-round, representative and benchmarkable to support both enterprise-wide and domain-specific learning goals and tasks in enterprise data science. Table2Vec significantly outperforms the existing shallow, boosting and deep learning methods typically used for enterprise analytics. We further discuss the research opportunities, directions and applications of automated universal enterprise representation and learning and the learned enterprise data DNA for automated, all-purpose, whole-of-enterprise and ethical machine learning and data science.

## Introduction

Data science plays a pivotal scientific role in transforming enterprise innovation and thinking[1–3]. Enterprise data science often involves a handful of heterogeneous relational and non-relational databases, enterprise data warehouses (EDWs) or marts, and other repositories from third parties or external resources. A common practice in data warehouse-driven business intelligence (BI) is to create an 'enterprise big table' consisting of domain-specific transactions and master data and informational and operational metadata about customers, businesses and their consumption of products and services from the enterprise[4,5]. Such a mixed deep-and-wide big table is then manipulated by specific business analysts and units to understand and analyze their domain subjects of interest and tasks, e.g., related to customers, products, services, productivity, marketing, competition, new product promotion, and interactions with customers. This *subject-oriented enterprise analytics* has been on the standard agenda of many enterprises for their data strategies and enterprise data science, becoming increasingly adopted to achieve better business performance, transformation, and decision-making.

However, such subject-oriented BI practice promoted by multinational vendors and implemented in numerous enterprises face some critical design limitations and technical and functional shortage. A fundamental issue is that no 'whole-of-enterprise data science' thinking, objectives, tasks and techniques are available to connect all-round but siloed businesses, processes, systems and data, as enterprise BI was deemed to address. Instead, most enterprise analytics actually 'decouples' enterprise businesses and data to specific domains and subject-specific analytical objectives, tasks and data, which are then addressed by analysts at their own business units[6,7]. This essentially shrinks from the global ambition of enterprise BI of connecting, fusing and bridging all-round siloed businesses and sliced data for whole-of-enterprise purposes but rather maintains the existing business silos and even creates new 'analytical silos'. Consequently, the existing approaches and results only capture partial enterprise intelligence, domain intelligence, human intelligence, and data intelligence[1], serve partial purposes and do not address enterprise-wide goals and challenges well. Unfortunately, this scenario also applies to related areas including cloud analytics and distributed and federated learning, where analytical silos remain on specific learning objectives, businesses, tasks and data by customized models[8].

In this work, we address this fundamental issue in the present best practice of enterprise BI and data science and suggest a *whole-of-enterprise* thinking with automated universal representation learning of all-round enterprise DNA. The *enterprise DNA* connects (1) domain-oriented transactions and master data, informational and operational metadata, and relevant external data; (2) customer demographics, activities, transactions, behaviors, interactions and communications with the enterprise; and (3) businesses including processes, status, consumption and feedback of its products, services, production, marketing, operations, and management, etc. The *automated universal representation learning* aims to fuse siloed businesses and data, resolve the business and data complexities in their representation fusion and address the subject-oriented limitations and their resultant analytical silos. It further enables *universal representation-oriented automated enterprise data science* for *universal enterprise representation and learning*, where the universal representation learning characterizes the whole-of-enterprise DNA about businesses and customers and expresses universal, representative and benchmarkable data genomes for enterprise-wide and domain-specific data science. The *automated universal enterprise representation and learning* further (1) automatically analyzes and manages data characteristics and quality issues in enterprise data; (2) generates universal representations on the all-round enterprise DNA to express the core data-genetic information about entities such as customers by connecting to their relevant businesses, processes and data in an enterprise; (3) conducts downstream data science and machine learning tasks; and (4) provides explainable and actionable results for enterprise-wide and domain-specific decision-making. Below, we summarize the methodologies, frameworks and their issues and opportunities of subject-oriented vs. universal representation-based enterprise data science and the data complexities in enterprise big tables and their learning challenges. We then introduce the concepts, rationale and designs of converting all-round enterprise DNA to their 'universal enterprise representations' on enterprise businesses and data. We illustrate automated universal enterprise representation and learning using a neural encoder Table2Vec to learn universal customer vector representations on commonly used heterogeneous big tables from EDW and evaluate the learned representations and results on both private and public big data against mainstream shallow, boosting and deep learning methods. Lastly, we expand the discussion to gaps, opportunities and applications of automated universal enterprise representation and learning for future research and innovation.

## Issues in Existing Enterprise Data Science

Here, we summarize the common framework and issues of the existing subject-oriented enterprise data science practice. We then discuss the gaps in the related work on learning from enterprise data for 'whole-of-enterprise' learning objectives and tasks.

### Subject-oriented Enterprise Data Science

In general, enterprise data science involves two sets of tasks: *data processing* to get data ready and statistically characterized for later management and reporting; and *data analytics* to make sense of data for decision-support[9,10]. Accordingly, an enterprise data science infrastructure connects multiple heterogeneous data sources, typically organized by heterogeneous relational databases[11], from business units and domains in an enterprise. For example, a major bank may have multiple relational databases and other storage and management systems to capture, store and manage data about their customer demographics and behaviors; products such as home loans, retail banking with saving and cheque accounts, online banking and credit cards, business banking, and insurance, etc.; and customer communications with the bank and for customer relationship management, etc.

To get the data ready and make sense of the data, business goals are defined per specific business domains and their requirements, the data scientists (forming data science teams in large enterprises) who are assigned with the business goals in an enterprise then acquire and gather the specific data relevant to the goals, process the data per their subject matter and task requirements, and then make the extracted data ready after cleaning, transformation, and initial pre-processing and statistical analysis[12]. Modeling techniques are then applied or developed to address the goals on the data. This forms the typical subject-oriented enterprise data science, commonly seen in most enterprises, as shown in Figure 1. On top of enterprise resources, each analyst or their team creates their own data sets (data marts) and data processing workflows, and applies or develops models to conduct specific learning tasks. Consequently, to cater for various enterprise analytics tasks, data analysts create many parallel data processing and modeling processes as shown in Figure 1.

The above data science exercise has been taken as a general practice by enterprise data scientists and forms a subject-oriented data science model. We summarize the process of this subject-oriented enterprise data science by specific analysts:

- Goal setting: set a business goal specific to a business domain and their problems and decision requirements, typically forming specific analytical subjects;
- Data ETL: exploiting highly task-specific data extracting, transforming and loading (ETL) methods to retrieve and collect the relevant data from one to several relevant sources;
- Data preparation: getting the extracted data ready for later learning by enhancing its data quality such as cleaning noises and imputing missing values, etc.;
- Feature engineering: selecting, constructing or mining for handcrafted features from the enhanced data; and
- Modeling: developing or applying analytical methods to conduct the learning tasks with results expected to address the specific business goals.

Such subject-oriented enterprise analytics is widely seen in many enterprises[13,14], enabled by off-the-shelf tools and solutions from multinational vendors. They are effective and efficient in terms of tailoring data, features and processes for specific learning tasks and business needs, enabling personalized processing and modeling by taking advantage of individual analyst's motivation, flexibility, knowledge and experience. However, they also incur various issues and limitations to more ambitious enterprise-wide and all-round data science, including (1) creating benchmarking gaps as the individually built data marts, processing workflows and models are hard to compare and benchmark; (2) increasingly generating data and analytical silos in an enterprise as an increasing number of closed-form specific data sets, processes and models are created; and (3) preventing transparent co-learning and team innovation, sharing and collaboration in the enterprise. Such subject-oriented and task-, analyst- and problem-specific practice is also highly time-consuming, nontransparent, irreproducible and unsustainable, failing to serve the whole-of-enterprise data science aims and needs[10].

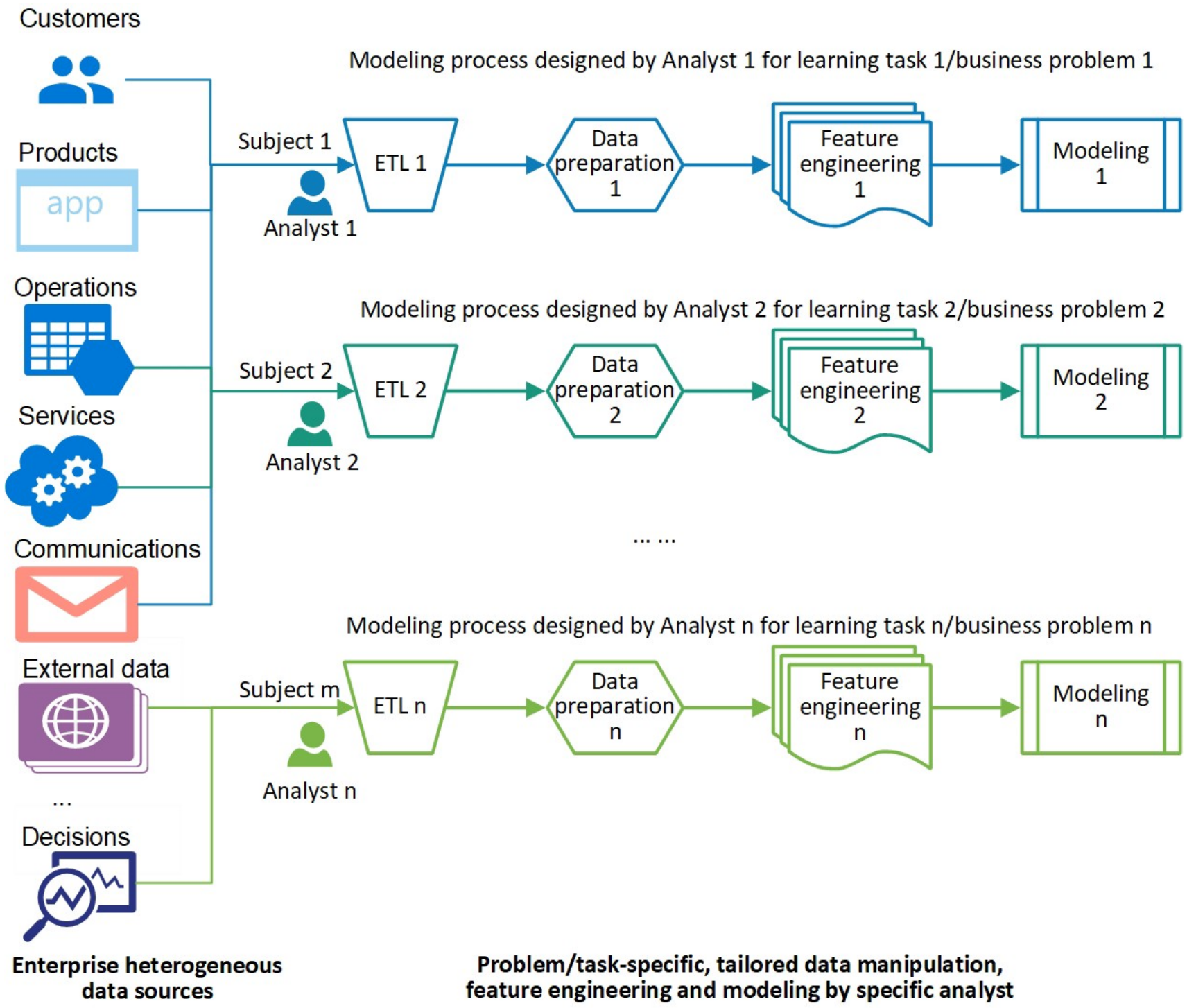

**Figure 1.** Subject-oriented Enterprise Data Science Tailored for Domain-specific Subjects, Problems and Tasks by Individual Analysts in an Enterprise. Driven by analytical subjects, analysts build workflows for each subject to implement data ETL, data preparation, feature engineering, and modeling.

## Gap Analysis of Related Techniques

The issues and limitations of subject-oriented enterprise data science are particularly caused by their problem-, task- and analyst-specific thinking, settings, methodologies, technical limitations and practices in the data manipulation and analytical processes. They are also owing to the limited design and capability of existing enterprise analytics systems and methodologies, which typically favor or support the following practical fashions and approaches.

- Each subject-specific learning task corresponds to a single purpose-driven analytical process, repeating the procedures from data extraction, transformation, manipulation, and feature engineering to modeling. Each task only involves partial data, features and potential of enterprise data, which are highly manipulated to fit the specific learning objective, model and process.
- Each data scientist independently and repetitively manipulates the same data sources and handcrafts own features for their business and task-specific analytics built by individual analysts or for specific domain/unit problem-solving purposes. As a result, an increasing number of isolated parallel data processing and modeling workflows, datasets, procedures and algorithms are generated on the data and in the enterprise, with more coming over time. This creates an increasing number of business, data and model slices in enterprise data science, forming analytical silos.
- The results delivered by a data scientist or from a data science team in an enterprise are often not benchmarkable across data, business units, and teams. It is often difficult or even impossible to evaluate, generalize, deploy and execute them against each other consistently and uniformly across the enterprise. The specific methods and results cannot be used for common benchmarking and general-purpose expansion, and do not mention enterprise-wide business deployment, production, problem-solving and the expansions of the results.
- As analytics is essentially done individually, none of the shareable and quality whole-of-enterprise, customer-oriented features are available in an enterprise data science system. Everyone has to customize something from scratch, forming a major barrier for general-purpose and whole-of-enterprise data science and cost-effective knowledge sharing and innovation.

Further, the subject-oriented enterprise analytics cannot well tackle the data characteristics and complexities of enterprise big data, as discussed in the section on challenges in learning from enterprise data, and fulfil the motivation and goals of supporting universal representations of all-round enterprise DNA and enabling whole-of-enterprise learning, as discussed in the section on quantifying all-round enterprise DNA. There are also significant gaps and issues in the machine learning algorithms available in enterprise BI systems and the literature to address these motivation and goals. They are also more suitable for siloed data and problem-, task- and analyst-specific practice. For example, many organizations have applied the following classifiers: k-nearest neighboring classifier (KNN)[15], naive Bayes[16], support vector machine (SVM)[17], decision tree[18], random forest[19], extreme gradient boosting (XGBoost)[20], and deep neural network (DNN)[21,22], etc. Although these individual methods are self-contained and easy to use, each of them has to be customized for specific learning tasks and repeatedly addresses the underlying characteristics and complexities and data quality issues widely seen in enterprise data, such as class imbalance[23], mixed data types[24], distribution heterogeneity[25], value sparsity[26], inconsistency[27], dynamics[28], and other data quality issues[10]. Table 1 illustrates the fitness of each of the above methods in addressing these data characteristics. These methods are typically applied by following the subject-oriented process as illustrated in Figure 1, where a specific workflow is customized for each task. This may create two dilemmas: one is that every workflow has to handle similar data issues and encode the data repetitively before specific learning is doable; the other is that it would be too costly or even impossible to enable each workflow to be fully capable or effective in addressing all data challenges and representations before the specific analytics is doable. This is why the so-called *80/20 data science dilemma*[1] is widely discussed in the communities. This 80/20 data science dilemma also causes deep problems, including duplicate and over manipulation of similar data sources, significant waste of valuable data and human resources on low-level jobs for data processing and engineering, and significant dilution of a more strategic and valuable commitment to the discovery of deep insights and value from data through creating cutting-edge and more powerful enterprise data science architectures and analytical models.

The above discussion briefly illustrates the strategic need for building whole-of-enterprise data genomes and learning universal enterprise representations that can (1) capture comprehensive customer information and interactions with its service providers and tackles various data characteristics automatically, shareable for any learning workflows, which can release the data scientists from handling the low-level intensive data manipulation; (2) create a universal representation over the cleaned all-round enterprise data, usable for any further learning tasks, which can serve as a shareable 'representation library' (to replace 'feature library') for benchmarkable learning, save costs in duplicated representation, and avoid incomparable representation learning by individuals; and (3) support end-to-end, closed-form and automated enterprise data science to reduce over handcrafting and enable the automated discovery of valuable findings.

[1] 80% of time is spent on preparing and getting the data ready for analytics, while only 20% of effort is made on the actual learning.

**Table 1.** Fitness of Self-contained Classifiers to Address Characteristics and Issues in Enterprise Data. Typical classifiers only focus on classification, requiring heavy and duplicated commitments by each analyst to address similar data quality issues. Table2Vec instead addresses both data quality issues and representation learning in one go, enabling end-to-end and automated enterprise data science.

| Classifiers | Imbalance | Mixed features | Heterogeneity | Sparsity | Inconsistency | Dynamics | Data quality issues |
|---|---|---|---|---|---|---|---|
| KNN | ✓ | | | | | | |
| Naive Byes | ✓ | ✓ | | | | | |
| SVM | ✓ | | ✓ | ✓ | | | |
| Decision Tree | ✓ | ✓ | ✓ | ✓ | | | |
| Random Forest | ✓ | ✓ | ✓ | ✓ | | | |
| XGBoost | ✓ | ✓ | ✓ | ✓ | | | |
| DNN | | ✓ | ✓ | ✓ | | | |
| Table2Vec | ✓ | ✓ | ✓ | ✓ | ✓ | ✓ | ✓ |

## Automated Universal Representation-based Enterprise Data Science

Here, we introduce our proposed automated universal enterprise representation and learning for benchmarkable and explainable enterprise data science. We first discuss the challenges in learning from complex enterprise data. We then introduce the quantification of all-round enterprise DNA, which characterizes and learns 'genetic elements' of an enterprise including their businesses, customers, activities, operations and data as universal representations of entities such as customers as enterprise data genomes. Lastly, we introduce a framework for learning universal representations with automated data augmentation on enterprise data.

### Challenges in Learning from Enterprise Data

Typically, enterprise data comes from enterprise data warehousing, where 'enterprise big tables' are constructed for analytical subjects by pulling mixed information from all relational and non-relational data sources and business lines and organized for customers[29,30]. Such big tables are indexed per all customers of the enterprise involving different products, services, processes, and communications with the enterprise. They have been widely used for business intelligence reporting and analytics in almost all enterprises with business intelligence systems[31,32]. However, they are not ideally designed for end-to-end and automated enterprise data science, and suffer from various issues preventing general-purpose[33] 'whole-of-enterprise' analytics. They involve complex data characteristics and pose various analytical challenges. Typical data characteristics and modeling complexities[1,10,27] in enterprise data science include:

- Inconsistency: The data describes customers who utilize different products, channels, services or processes of the company; customers have different demographics, conditions, profiles, behaviours, communications with the enterprise, and value to the enterprise, etc. Records are simply merged into one table per customers without a structure to present the characteristics (e.g., behavioral and interactive relations) in their physical worlds.
- Heterogeneity: Mixed data types exist in the data, including categorical and numerical features, and others such as time-related, imagery and textual data, etc. Each has different formats, distributions, structures, etc. The table is a mixture of everything without heterogeneity mitigation.
- Relations, interactions and couplings: Heterogeneous features from different business units of an enterprise for a customer may be explicitly and implicitly, strongly or weakly, and locally or globally related and coupled with each other in terms of various reasons, factors, and opportunities. Customers may interact with each other or with service providers through different communication channels and may be more or less coupled, e.g., from the same domain or share similar preferences and feedback.
- Ultrahigh dimensionality: The big table usually consists of more than thousands of features associated with each customer, which present highly inconsistent characteristics (e.g., distributional balance and sparsity).
- Dynamics: Some features are dynamic, while others are static; customers and their circumstances, service usage and interactions with the enterprise change over the time; new customers may join while some existing ones may churn. Many features may present evolving to nonstationary characteristics.
- Labeling inconsistency: Some customers are labelled with categorization by the enterprise, e.g., typically labeled as 0/1, showing whether a customer is likely to churn or has committed fraud; while most transactions are affiliated with no labels, which does not mean they are not categorized. In the table, all digitized labels are often mixed with the customers without distinguishable label semantics on a customer, e.g., whether a customer with label '1' refers to churn in one case while fraud for another.

- Data quality issues: There are often overwhelming data quality issues in such big tables, e.g., bias, error, redundancy, sparsity, imbalance, noise, and completeness (missing values). Many customers have very limited to no transactions affiliated with them, while others may be associated with intensive records.
- Imbalance and irregularity: As a big table is a mixture of everything, one customer's transactions likely differ from others in terms of data availability, quantity, feature value distributions, feature and customer couplings, and dynamics, etc. Some customers may subscribe to more products and services than others. A few customers may be associated with concerning categorizations such as fraudulent behaviors or potential attrition.
- Hierarchy: Hierarchy naturally exists in an enterprise's organisational structures, businesses and processes, which is also reflected in their data. In addition, customers may be clustered into groups and form hierarchical categories, the same as features forming feature subspaces or domain-specific feature sets.

In addition, there may be other data, behavior, business and decision-related characteristics and complexities[1] embedded in enterprises and their data. They together make it highly challenging to analyze the data directly by existing machine learning and data analytics systems and also challenge classic analytical methodologies[34], data integration strategies including master data management[35,36], and cloud-based application integration.

## Quantifying All-round Enterprise DNA

Accordingly, to enable shareable and benchmarkable enterprise data science and address the above need and issues, a fundamental task is to quantify the all-round enterprise DNA about businesses and customers in an enterprise as universal and representative enterprise representations. *Universal enterprise representations* capture the explicit and implicit multifaceted aspects of every customer and his demographics, activities, circumstances of service consumption, and interactions affiliated with all relevant businesses in an enterprise. They thus capture the data genomes of each customer connected to the enterprise and can also serve both general-purpose and specific analytical and decision-support objectives. Such a universal representation addresses the following objectives.

- The business and information related to each customer in an enterprise can be encoded by a universal vector representation, and the representations of all customers are consistent, comparable and benchmarkable. These *whole-of-customer representations* capture all-round customer DNA from the entire enterprise.
- The representations are built on all businesses, including all of business products, services, problems, tasks and interactions between a customer and the enterprise, to enable *whole-of-business customer representations*.
- The customer representations are built on all sources of enterprise data related to customers, including multi-relational and non-relational and internal and external resources, to enable *whole-of-data representations*.
- A customer representation is built on all features in enterprise data that are heterogeneous, sparse, redundant, inconsistent, coupled and dynamic and are associated with various data quality issues. It supports inbuilt, automated and optimal feature engineering of all relevant features for both general and specific learning tasks, enabling *data-augmented automated customer representations*.
- Such a representation captures all aspects of important and relevant information and features about a customer and the enterprise, which forms the 'data DNA'[1] of the enterprise and its customers. Similar to the roles of a biological organism's gene and DNA, an enterprise's customer representation serves as a representative, benchmarkable and universal data genome of a customer, to produce *enterprise customer data DNA*, usable for the understanding and processing of different business problems and learning tasks relevant to the customer and enterprise.
- The representation can enable better data science performance, transparency, reproducibility, shareability, result interpretability and benchmarking, modeling transferability, and digital production and automation.

Accordingly, Figure 2 illustrates the motivation for capturing all-round enterprise DNA and generating the universal data genomes of customers. This addresses the aforementioned issues, limitations and challenges in existing enterprise data science and results in the better practice of enterprise data science. Such learned enterprise customer data genomes capture the 'DNA and genes' of all customers and businesses, including (1) the features describing the demographics, behaviors, service consumption, interactions, dynamics and feedback of each customer; (2) the features capturing the enterprise's profile, behaviors, assets, governance, dynamics and decisions that may involve various business units including the operations, design, production, marketing, sales and services of products; and (3) various stages of a customer's interactions with an enterprise, including inquiries, registrations, consumption and feedback on enterprise products and services and possible transition and change during engagement.

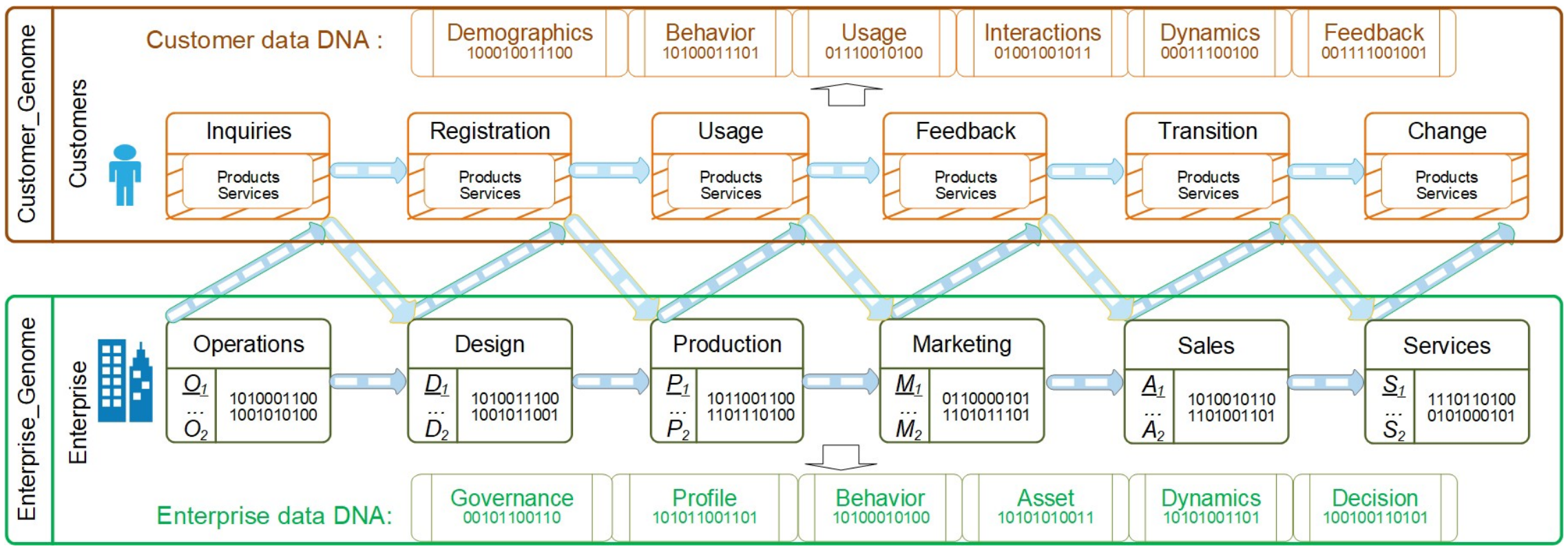

**Figure 2.** Learning Enterprise Representations to Capture the Universal, Representative and Benchmarkable Data Genomes of an Enterprise and Their Customers: Enterprise Data DNA, and Customer Data DNA.

### Automated Universal Enterprise Representation and Learning

To enable enterprise-wide benchmarkable data science building on EDW, all the relevant data are gathered in whole-of-business enterprise big tables for all customers or data marts with those customers sitting in one business domain. Since such enterprise big tables are massive, high-dimensional, heterogeneous and usually associated with various quality issues, developing a universal customer representation of each customer on the entire complex enterprise data is thus highly desired yet challenging.

While there may be different methods to capture and represent enterprise and customer data DNAs in Figure 2, here we introduce a learning approach to learn the universal representations of all customers which captures customer information and their activities and interactions with an enterprise by involving the enterprise products, activities and processes and communications with customers[37]. The learned universal customer representations can play critical and strategic roles for enterprise data science, including serving the general and specific purposes of enterprise learning, business understanding, and decision-support and making the further learning cost-effective, easy-to-implement, and benchmarkable[38].

Of the different ways to achieve the above purpose[39,40], Figure 3 illustrates the workflow of our proposed Table2Vec[2] that restructures the one in Figure 1 toward representing the whole-of-enterprise data DNA for various business objectives and learning tasks. We illustrate the universal enterprise learning of enterprise and customer DNA through building a Table2Vec deep neural network to capture enterprise-wide customer data genomes in enterprise big tables extracted from EDW. With multi-relational data sources in an enterprise (e.g., a bank as an example), customer data in different tables are extracted, transformed and loaded (ETL) from multiple heterogeneous data sources into an enterprise big table. Table2Vec then automatically identifies and analyzes different data characteristics in the big table and generates a universal customer representation on all data relevant to the customer. The resultant customer representations of all customers are universal, shareable and reusable for different learning tasks. Data scientists with different learning tasks can further manipulate the representations and build task-specific models or problem-specific applications. In addition, interpretation techniques[42] can be developed to provide the interpretability of the unified customer representations, e.g., characterizing data genomes incorporated in the final representations and mapping the learned representation to observable features to identify major factors driving and explaining various business outcomes and learning results. Consequently, Table2Vec addresses the aforementioned enterprise data science objectives, data characteristics, modeling complexities and the motivation of learning enterprise data genomes; tackles the limitations in existing enterprise data science; and serves as a universal and strategic enterprise customer's genetic representation for various purposes and tasks in enterprise data science.

[2]Interested readers may also refer to[41] where we introduce the Mix2Vec neural encoder designed together with Table2Vec but focusing on representing mixed data.

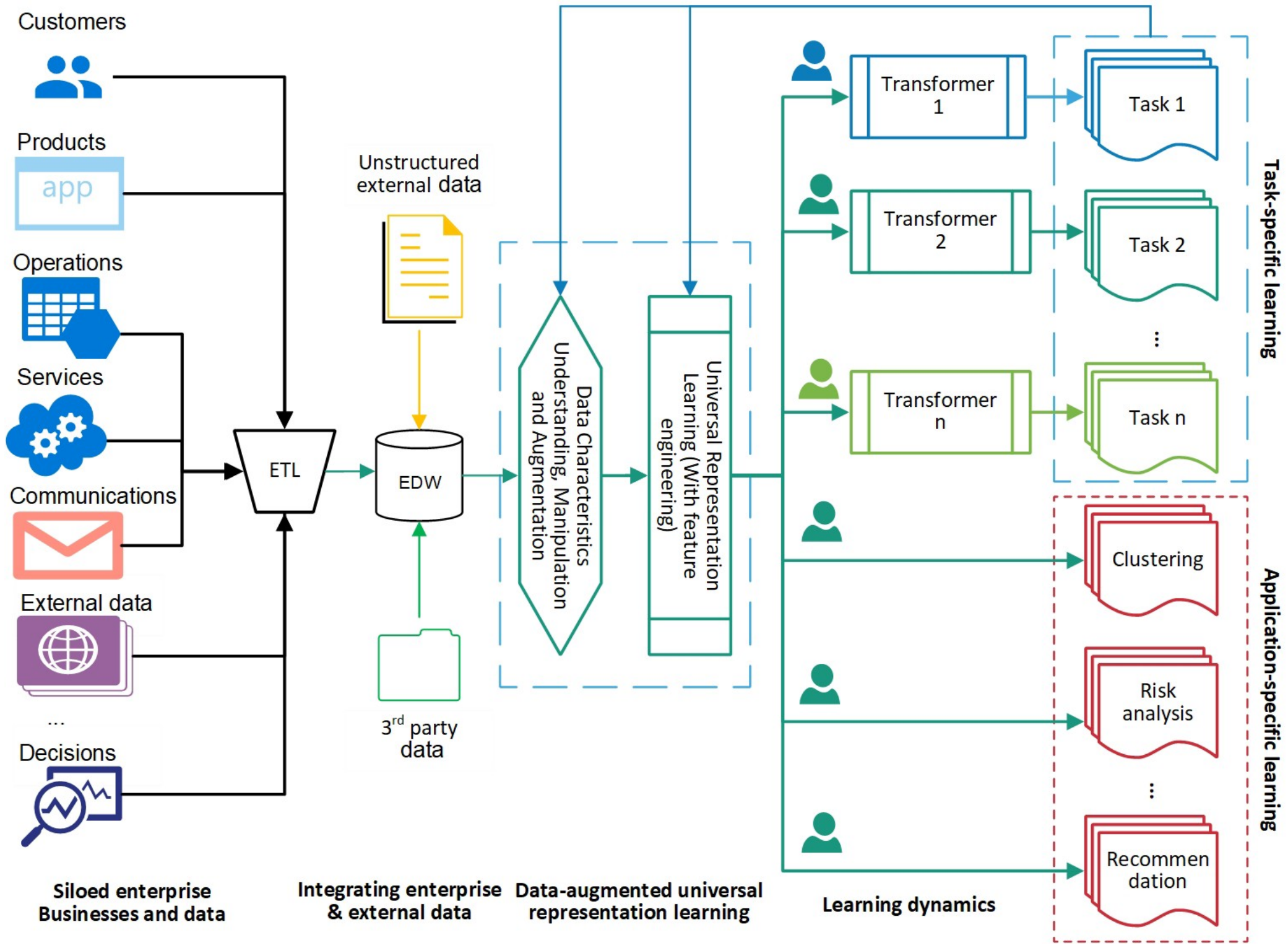

**Figure 3.** Automated Universal Representation-based Enterprise Data Science for General-purpose and Specific Tasks in an Enterprise. Data-augmented universal representations are learned on integrated enterprise data and enhanced to address business and data characteristics and dynamics, which can then be used for enterprise-wide and domain-specific learning tasks or applications in the enterprise.

## Table2Vec: Interpretable Automated Enterprise Representation and Learning with Augmentation

### The Table2Vec Neural Networks

The Table2Vec representation learning addresses the complexities in multi-relational enterprise big tables and generates general-purpose customer representations with the following design thinking and working mechanisms.

- Data quality mitigation: handle the bias and impact caused by enterprise data quality issues including sparsity, inconsistency and missingness on skewing the learned customer representations toward biased results.
- Modelling dynamics: capture the dynamics of business, data and customers over time.
- Multi-task learning: learn universal customer representations to fit multiple learning tasks and various business objectives.
- Enabling interpretability: learn task-sensitive customer patterns and their discriminative features in the representations that can explain customer behaviors and outcomes.

Accordingly, Figure 4 shows the framework of the neural Table2Vec model with automated data characteristics analysis and augmentation. Table2Vec has various functional modules to automatically analyze the data heterogeneity and dynamics, embed heterogeneous features, and unify the respective representations. Figure 5 further illustrates the workflow from an input big table to a universal representation of each customer by the Table2Vec model. Table2Vec learns each customer's representation as a low-dimensional vector from an enterprise big table which consists of customer records with mixed types of heterogeneous multi-relational data. Table2Vec first recognizes the types of futures in the table and differentiates numerical features from categorical and time (date) features by NC-Recognizer. SD-Recognizer further distinguishes static and dynamic data by detecting the change of numerical and categorical data.

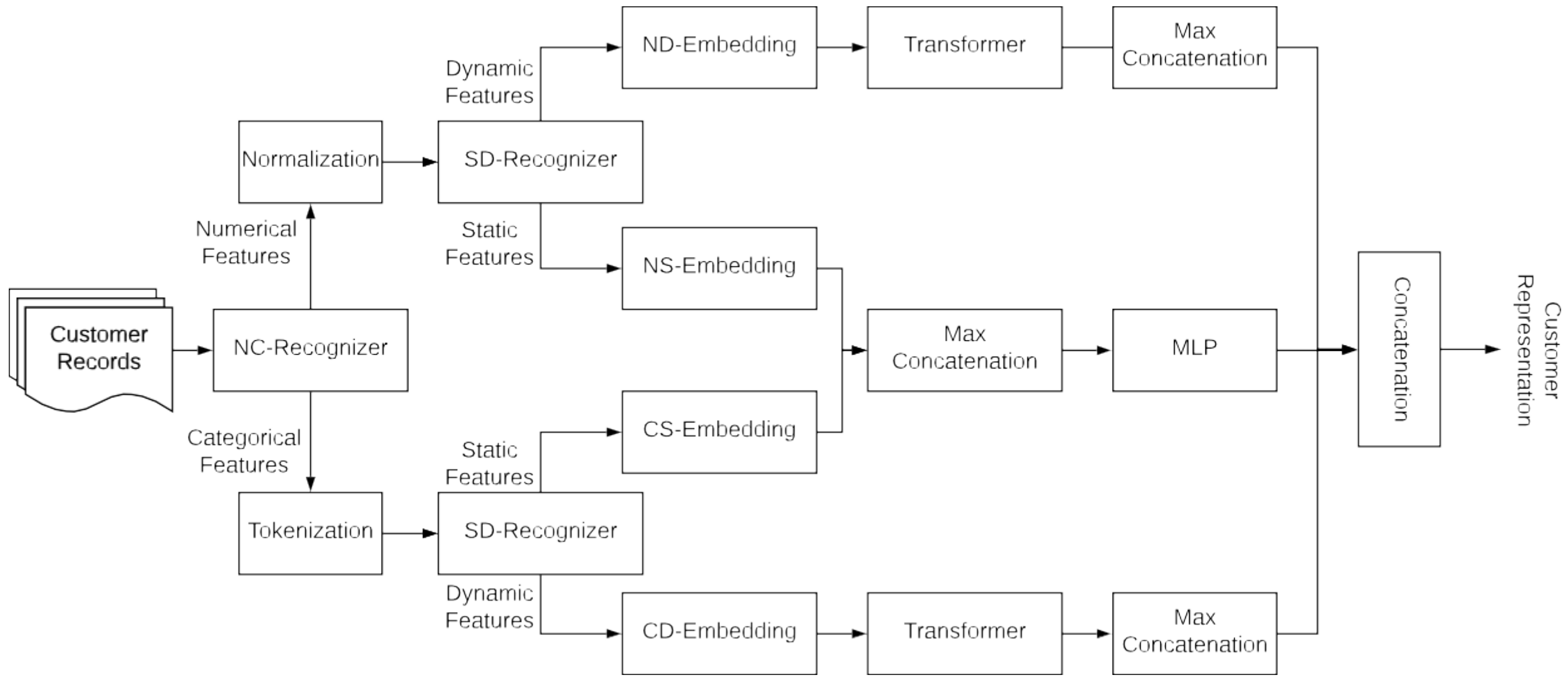

**Figure 4.** The Neural Framework of Table2Vec with Automated Data Characteristics Analysis and Augmentation. NC-Recognizer: automatically recognize and distinguish numerical and categorical features; SD-Recognizer: automatically recognize and distinguish static and dynamic features; ND-Embedding: vectorize dynamic numerical feature values; NS-Embedding: vectorize static numerical feature values; CS-Embedding: vectorize static categorical feature values; and CD-Embedding: vectorize dynamic categorical feature values.

Accordingly, mixed features in a given big table are split into four types of features: static categorical features, static numerical features, dynamic categorical features, and dynamic categorical features. Further, Table2Vec vectorizes these four types of recognized features by respective embedding modules: CS-Embedding vectorizes the static categorical feature values, NS-Embedding vectorizes the static numerical feature values, CD-Embedding vectorizes the dynamic categorical feature values, and ND-Embedding vectorizes the dynamic numerical feature values, respectively. In addition, to capture the dynamics and valuable sequential patterns in customer behaviors, Table2Vec adopts the universal transformer[43] to model their dynamic categorical and numerical embeddings. To resolve data sparsity and inconsistency, Table2Vec uses a max concatenation layer to leverage the most significant patterns from a huge number of features in each feature type. Lastly, Table2Vec concatenates and transforms the embedded vectors learned from the above four types of features by fully connected layers to form the unified customer representations.

To generate general-purpose customer representations, Table2Vec adopts two mechanisms to train neural networks. The first mechanism is to use the Table2Vec representations to reconstruct the multiple random mappings of the original tables. This guarantees the learned representation embeds the noise-tolerant essential information contained in the original table from different sources as much as possible. The second mechanism is to conduct multiple relevant learning tasks to regularize the model learning. Specifically, Table2Vec transforms the general-purpose customer representation to a task-specific representation by fully connected layers for each task. For all learning tasks, Table2Vec jointly minimizes all task-specific loss functions based on the task-specific representations to regularize the training of its neural networks. Consequently, the resultant customer representation embeds all information required by different tasks to serve the general-purpose objectives discussed in the section on universal representation-based enterprise data science, which can be used by various learning tasks, e.g., customer profiling, customer similarity analysis, and the next-best conversation recommendation. The learned customer representation can be further used for specific learning tasks by respective fully connected dense layers. In this work, we illustrate Table2Vec with the customer representations trained in a supervised way for either one task learning or multi-task learning, where classifiers are given with labels.

## Heterogeneous Feature Recognition and Embedding

Given enterprise data D with customer set $U$ and feature set $F$, Table2Vec generates universal customer representations $\mathbf{E}$. It automatically recognizes heterogeneous features and processes them accordingly, in addition to a user interface provided to allow analysts to assign types to features. Two feature recognizers automatically recognize and distinguish numerical and categorical features (i.e., by the NC-Recognizer) and static and dynamic features (i.e., by the SD-Recognizer), respectively. NC-Recognizer conducts the following tasks and workflow: 1) recognizing the types (e.g., object, float, integer, and date) of all input features; 2) treating objects as categorical, floating features as numerical, and date-related features as a datetime index; and 3) generating the statistics of the number of unique values of integer features and identifying a numerical feature if its count is larger than a prefixed threshold, otherwise a categorical feature. SD-Recognizer identifies dynamic features by obtaining feature dynamics per customer over time. Specifically, we first order the transactions affiliated with each customer chronologically, and then count the statistic change of successive feature values per customer as follows:

$$D_{,f} = \begin{cases} Count(\mathbf{V}_{u,f}^{1:N} \neq \mathbf{V}_{u,f}^{0:N-1}), & if\ f\ is\ categorical, \\ |\mathbf{V}_{u,f}^{1:N} - \mathbf{V}_{u,f}^{0:N-1}|, & if\ f\ is\ numerical. \end{cases} \tag{1}$$

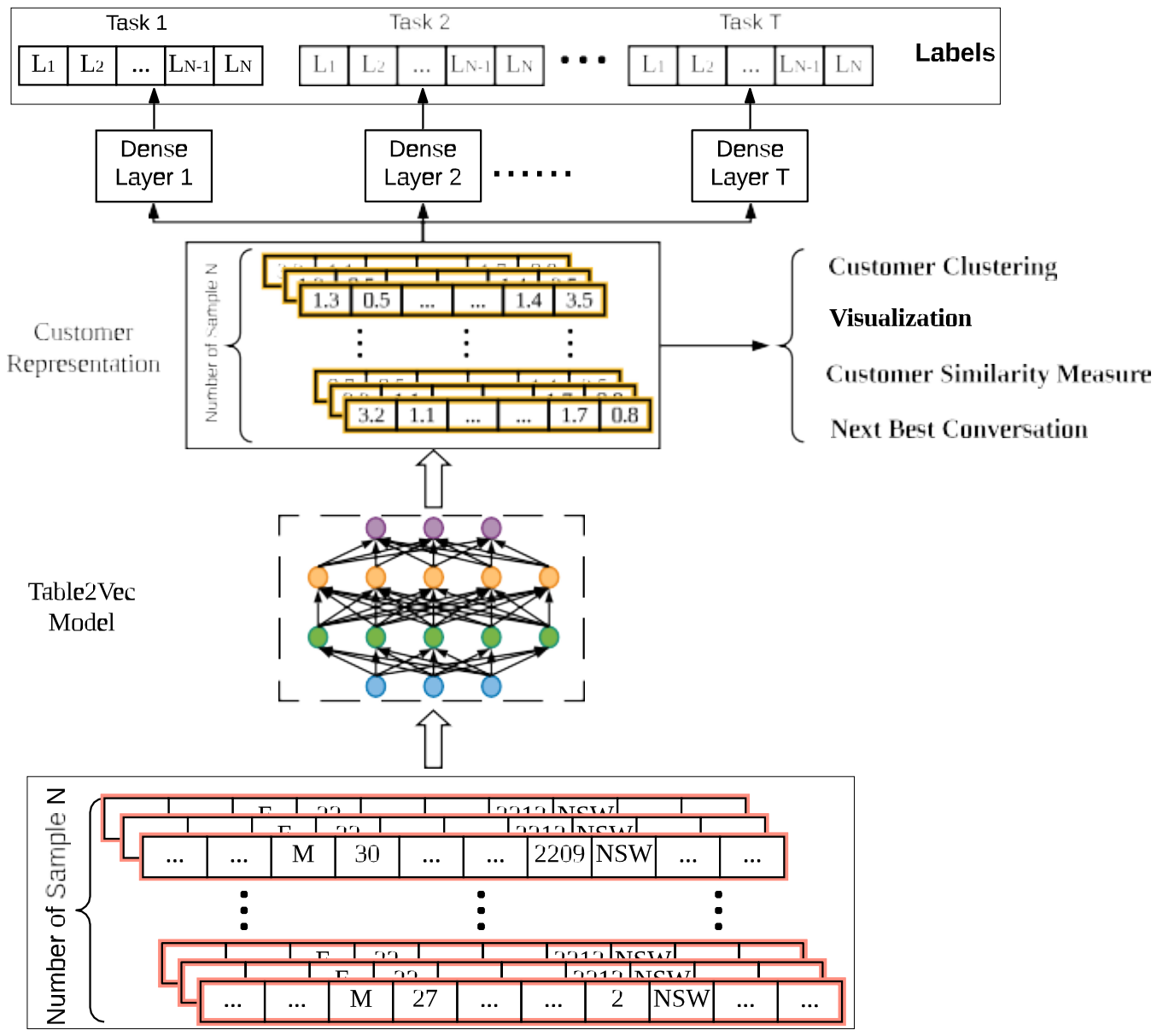

**Figure 5.** The Information Flow of Table2Vec. Customer-related records are extracted from the enterprise big data table, on which Table2Vec learns a representation of each customer; the learned customer representation can be used for business problem-specific applications or be further transformed to support learning tasks.

$\mathbf{V}_{u,f}$ denotes the feature vector of customer $u$ on feature $f$, and $Count(condition)$ calculates the count of the *condition* satisfied. Consequently, we obtain a difference matrix $\mathbf{D} \in \mathrm{R}^{|U| \times |F|}$, which has the statistics of feature dynamics on each customer. With the matrix, we first prefix the threshold $T_D$ to filter dynamic customer-feature pairs, i.e., $(u, f)$, indicating whether feature $f$ is dynamic in terms of customer $u$. Then, we count the user-feature pairs per feature to measure the statistics of how many customers show dynamics on each feature and use another threshold $T_F$ to filter the final dynamic features.

$$Type(f) = \begin{cases} dynamic, & if\ D_f > T_f, \\ static, & if\ D_f < T_f. \end{cases} \tag{2}$$

where $D_f = Count(\{D_{u,f} > T_D\})$.

Consequently, the NC-Recognizer automatically recognizes and labels categorical (such as gender, business product type, and customer feedback polarity) and numerical (e.g., home loan amount and duration) features. The SD-Recognizer recognizes and differentiates features with more significant dynamics. For example, a home loan type that receives a significant quantity of customer complaints would be regarded as more market-sensitive and dynamic than another type with few concerns, the former loan type would be treated as a dynamic feature while the latter would be treated as static. After distinguishing the feature types, we represent different types of features by their respective embedding methods, including dynamic categorical feature embedding by ND-Embedding for dynamic numerical feature values, NS-Embedding for static numerical feature values, CS-Embedding for static categorical feature values, and CD-Embedding for dynamic categorical feature values.

## Data Quality Augmentation

As discussed in the section on challenges in learning from enterprise data, enterprise data are associated with various quality issues including sparsity, inconsistency, and missing values. Data sparsity spreads from features to customers, bringing great challenges to learning general-purpose customer representations on not only the feature value embedding but also skewing the representation learning. A series of treatments including normalization, imputation, position-based numerical embedding, and max concatenation layers are implemented to make sparse features easy to handle, prevent sparse features from skewing representation learning, and effectively feed the extracted features to higher-level networks.

*Tokenization, normalization and imputation*. For categorical features, we tokenize (e.g., by unigrams, bigrams, or n-grams-based word embeddings) all of their feature values including the missing values, where we treat missing values as a specific kind of value for convenience. Differently, we apply a uniform normalization to scale numerical features, and then impute zeros for missing values. The uniform normalization is defined as follows:

$$N_{uni}(x) = \frac{x - min(X)}{max(X) - min(X)}, \ x \in X. \tag{3}$$

We convert non-missing values to the range of $(0, 1)$ and treat missing values as zeros, which benefits network training and relieves the influence of missing values.

*Position-based numerical embedding*. For categorical features such as product types and complaint categories, after using the embedding look-up method, we further obtain two embedding matrices, $E^u_{(CS}$ and $E^u_{(CD)}$, for each customer $u$. For numerical features such as customer credit limit and home loan amount, the general embedding is not effective since enterprise data is often associated with a high proportion of missing values. We propose a position-based numerical embedding for numerical features to relieve the influence of missing values[44]. The following embedding method is used to extract enough information from non-missing values. Specifically, we fix the input order of numerical features and then transform each numerical value to a dense or zero vector:

$$E^u_{i(N)} = v^u_i * \mathbf{M}_i \tag{4}$$

$$E^u_{(N)} = \mathbf{v}^u * \mathbf{M}. \tag{5}$$

where $E^u_{i(N)}$ is the value embedding for the $i^{th}$ numerical feature of user $u$ and $\mathbf{M}$ is the look-up matrix numerical embedding. Since we treat numerical static and numerical dynamic features separately, we apply the position-based numerical embedding to two kinds of features, respectively. Accordingly, we obtain two embedding matrices, $E^u_{(NS)}$ and $E^u_{(ND)}$, for the two kinds of numerical features of each customer $u$.

*Max concatenation layers*. To retain effective information and address the sparsity issue, we propose a max concatenation layer to aggregate the initial embedding output. For a given embedding matrix $E$, we generate an aggregated embedding vector by the maximum operator:

$$V^u = \{max(E_{:,1}), \cdots, max(E_{:,d})\} \tag{6}$$

where $max(E_{:,j})$ is the $j^{th}$ column of the matrix $E$, and $d$ denotes the embedding dimensionality.

## Modeling Business and Data Dynamics

Enterprise business and data evolve over time, business lines, products, and customers and involve diverse changes and dynamics. Table2Vec adopts the universal transformer[43] to model the dynamics of enterprise business and data. Given an embedding sequence $\mathbf{E}^0 \in R^{n_s \times n_e}$ of length $n_s$ and embedding dimension $n_e$, the universal transformer iteratively computes the representations $\mathbf{E}^t$ at step $t$ for all $n_s$ positions in parallel by applying the multi-headed dot-product self-attention mechanism[45], followed by a recurrent transition function. It also adds residual connections around each of these function blocks and applies dropout and layer normalization[45].

Formally, the universal transformer calculates $\mathbf{E}^{t+1}$ from $\mathbf{E}^t$ as follows,

$$\mathbf{E}^{t+1} = LN(\mathbf{A}^{t+1} + TS(\mathbf{A}^{t+1})), \tag{7}$$

where

$$\mathbf{A}^{t+1} = LN((\mathbf{E}^t + \mathbf{P}^{t+1}) + MHSA(\mathbf{E}^t + \mathbf{P}^{t+1})), \tag{8}$$

and $LN(\cdot)$ is the layer normalization defined in[46], $TS(\cdot)$ is a fully-connected neural network that consists of a single ReLU activation function between two affine transformations applied position-wise, $\mathbf{P}^{t+1} \in R^{n_s \times n_e}$ is two-dimensional (position, time) constant coordinate embeddings defined in[45], and $MHSA(\,)$ is the multi-headed dot-product self-attention that is discussed below.

With $k$ heads, the multi-headed dot-product self-attention is calculated as follows,

$$MHSA(\mathbf{E}^t) = [head_1, \cdots, head_k] \cdot \mathbf{W}^O, \tag{9}$$

where

$$\begin{aligned} head_i &= Attention(\mathbf{E}^t\mathbf{W}_i^Q, \mathbf{E}^t\mathbf{W}_i^K, \mathbf{E}^t\mathbf{W}_i^V) \\ &= softmax(\frac{\mathbf{E}^t\mathbf{W}_i^Q\mathbf{W}_i^{K\top}\mathbf{E}^{t\top}}{\sqrt{n_e}})\mathbf{E}^t, \end{aligned} \tag{10}$$

and $\mathbf{W}^Q, \mathbf{W}^K, \mathbf{W}^V \in R^{n_e \times n_e/k}$ and $\mathbf{W}^O \in R^{n_e \times n_e}$ are the network weights that need to be learned.

To adaptively control the iterative steps, the universal transformer introduces a halting mechanism on dynamic adaptive computational time (ACT)[47] to each position. The final output $\mathbf{E}^f \in R^{n_s \times n_e}$ of the universal transformer is a concatenation of the value of each position when the position is halted by ACT as follows,

$$\mathbf{E}^f = [\mathbf{E}_{1,\cdot}^{h_1\top}, \cdots, \mathbf{E}_{n_s,\cdot}^{h_{n_s}\top}]^\top. \tag{11}$$

Finally, Table2Vec calculates an embedding vector $\mathbf{e}^d \in R^{1\times n_e}$ as the embedding of dynamic features such as an evolving home loan amount from their universal transformer representations $\mathbf{E}^f$ as follows,

$$\mathbf{e}^d = [\mathbf{E}_{1,\cdot}^f, \cdots, \mathbf{E}_{n_e,\cdot}^f] \cdot \mathbf{W}^D, \tag{12}$$

where $\mathbf{W}^D \in R^{n_s n_e \times n_e}$ is a network weight matrix to be learned.

## Enabling Representation Interpretability

The Interpretation Module (IM) in Table2Vec maps each position in the Table2Vec representation to patterns existing in the original table, as shown in Figure 6. Inspired by[48], we implement the IM in three steps for each position in the Table2Vec representation. In the first step, for each position, IM finds the customers in the original table who are associated with the $k$-largest values in the position, namely $k$-sensitive customers. In the second step, for each identified customer, IM randomly masks a feature of the customer at a time-point and feeds the masked original information to Table2Vec to generate a new representation. In the third step, IM compares the values in the position of the generated new representation with that of the original representation and finds the changed features that make the value difference larger than a threshold, namely the sensitive features. Finally, IM interprets the patterns captured by the position as the values of its sensitive features describing its sensitive customers. The rationale is that sensitive customers must contain a pattern that is captured by the position, and the sensitive features further identify which features of the sensitive customers hold the pattern.

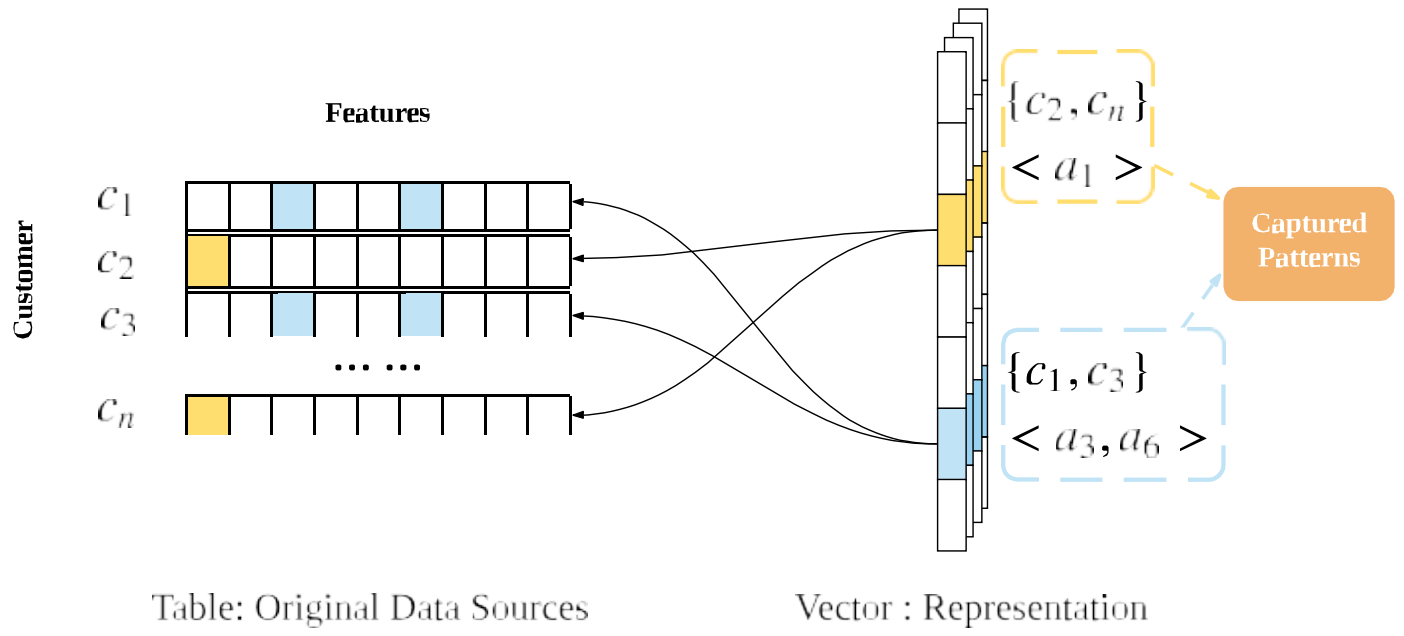

**Figure 6.** The Interpretation Module of Table2Vec. The features discriminating a customer's matter of concerne is traced from the original data, forming the customer's pattern corresponding to the learned discriminative representations.

Lastly, we summarize the Table2Vec processes of learning universal representations $\mathbf{E}$ of enterprise data D with customers $U$ and features $F$ automated data augmentation.

**Algorithm 1** *Table2Vec for universal enterprise representation learning*

**Input:** Heterogeneous enterprise data with customers $\{u\}$, features $\{f\}$ and values $\mathbf{V}_{u,f}$, as well as quality issues
**Output:** Universal customer representation $\mathbf{E}^{u}_{(\cdot)}$ over all features

Initialize a $|U| \times |F|$ enterprise big data table $D$ collecting all enterprise data
**for** $u$ in $U$ **do**
  **for** $f$ in $F$ **do**
    Recognize feature categories by NC-Recognizer and SD-Recognizer: $D_{u,f}(\mathbf{V})$ and $Type(f)$;
    Augment data quality: tokenization, normalization, and imputation, etc.
    Model business and data dynamics by Transformer: $\mathbf{E}^{f}$
    Concatenate feature-specific representations and generate the universal customer representation: $\mathbf{e}^{d}(f)$
    Generate representation interpretability $\sim (u, f)$
  **end for**
**end for**
**return** $\mathbf{E}$

## Results

The Table2Vec model was evaluated using both private and public data. Below, we show the results of applying it on the enterprise data from a major Australian bank and on a publicly available large Kaggle data set, respectively. Both data sets are imbalanced and have data quality issues. We briefly introduce the baseline methods, results, and their interpretation by addressing their respective data characteristics.

### Baseline Methods

We compare the Table2Vec-based neural classifier described in Figure 5 with the following popular shallow and deep classifiers based on the binary labels provided in the two sets of data. These baselines are chosen because (1) they are mostly used for enterprise data science; (2) ensembles like random forests, GBoost and XGBoost are typically applied for improving performance; (3) they cover different learning mechanisms including tree models, ensembles, distance learning, probabilistic learning, and deep learning; and (4) a state-of-the-art deep neural network (DNN) is also compared.

- *Decision Tree*: it is a common classification method with a tree structure, easy to implement and has a good interpretability.
- *Random Forests*: it is an ensemble of decision trees for classification, addressing the overfitting issue in single decision trees.
- *GBoost*: it is an ensemble of decision trees to allow the optimization of an arbitrarily differentiable loss function.
- *XGBoost*: it is a scalable, portable and distributed GBoost edition.
- *KNN*: it is short for $k$-nearest neighbors, a non-parametric distance-based classifier commonly used for classification and regression.
- *Naive Bayes*: it is a simple probabilistic classifier based on the Bayes' theorem and independence assumption.
- *DNN*: it is a state-of-the-art deep neural network widely used for supervised learning with demonstrated performance.

For a fair comparison, we well tune the parameters of the above baseline methods and report their best classification performance. We compare the Table2Vec-enabled classifiers with the above baseline models in terms of AUC, F-Score and weighted accuracy on the imbalanced data for fair and appropriate evaluation. All baselines were implemented and tested by us and then independently evaluated by the Bank's data science group to ensure the robustness, transparency and reproducibility of our designs and results.

### Results on Banking Customer Representation

The banking enterprise big table consists of $284,547$ records of $47,314$ customers collected from 30 April 2016 to 29 September 2017. Each customer record consists of $1,208$ features, describing various information, such as customer demographics and behaviors, product information, and customer interactions with different business lines and service channels in the bank. Part of these features evolve over time, resulting in more than six records for each customer collected in the data. In addition, all customer records were collected from two business lines - home loan and credit card - with inconsistent binary labels given to issues identified in the loan and credit card businesses by the bank. The data is complex as it consists of heterogeneous features[41], some of which are dynamic, all of them are sparse with a significant proportion (up to 5%) of features having missing values, and the labels provided by the bank are binary without semantic differentiation.

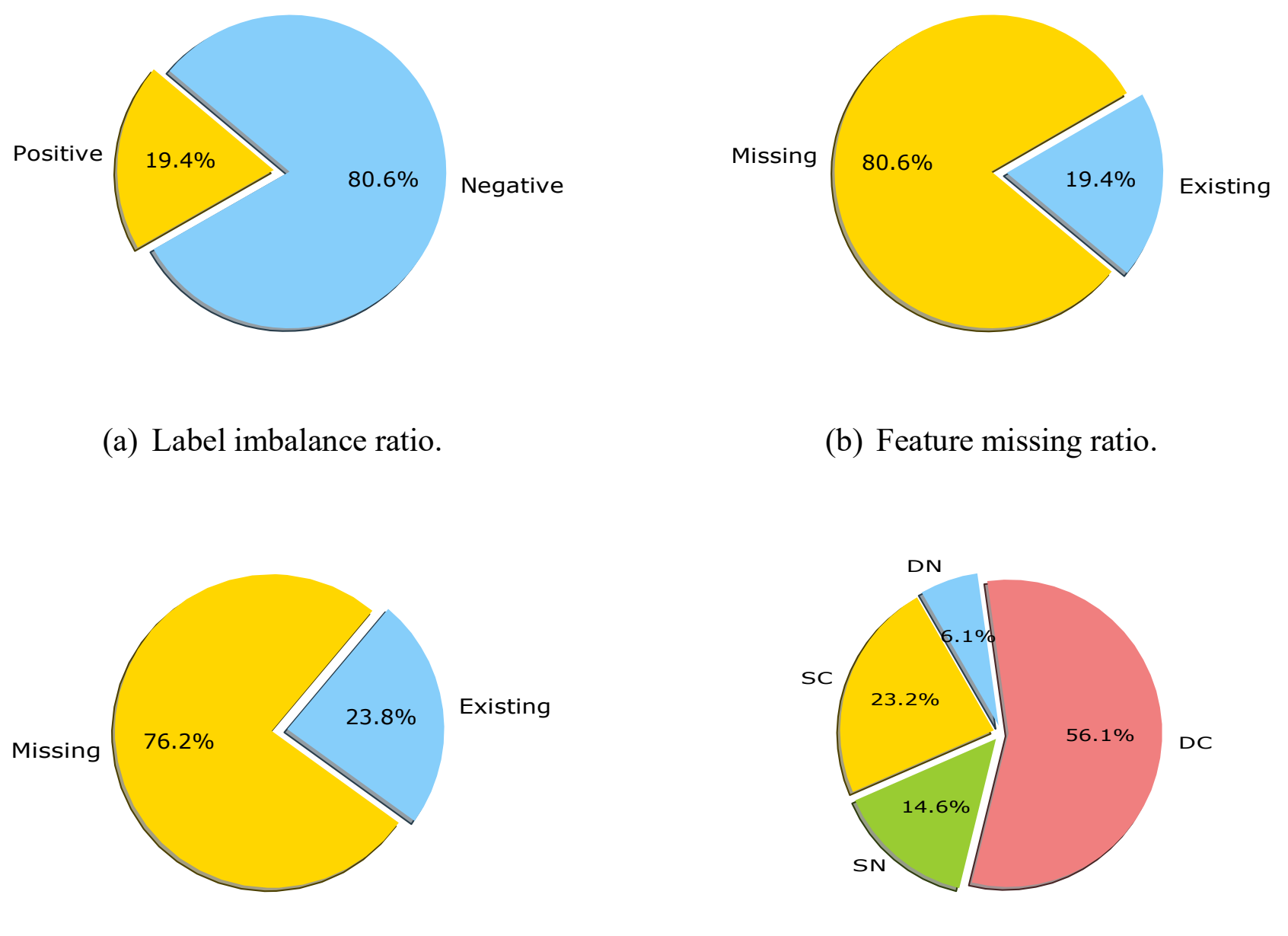

(a) Label imbalance ratio.

(b) Feature missing ratio.

(c) Feature structural missing ratio.

(d) Data type distribution.

**Figure 7.** Data Characteristics of the Enterprise Data from a Bank. SC refers to static categorical data, SN refers to static numerical data, DC refers to dynamic categorical data, and DN refers to dynamic numerical data.

For a better understanding of the complex data characteristics, we summarize several important statistics of the sampled big table in Figure 7. It shows that the ratio between positive and negative labels is nearly 1 : 4 and the ratios of the features with missing values and the features with structural missing values are both above 75%, evidencing the imbalance and sparse nature of enterprise big tables. Here, *structural missing* means that all values of a feature are missing in all records of a customer, indicating this customer may have nothing to do with the business indicated by the feature. Both *feature missing ratio* and *feature structural missing ratio* are calculated over the records of each customer and then averaged over all customers.

Furthermore, we apply the NC-Recognizer and SD-Recognizer modules in Table2Vec to automatically distinguish numerical (N), categorical (C), dynamic (D) and static (S) features. We thus obtain four feature combinations: SN - static numerical features, DN - dynamic numerical features, SC - static categorical features, and DC - dynamic categorical features, and their ratios in the data are shown in Figure 7(d). The results obtained in the sampled big table show that the enterprise big data is dynamic, inconsistent, high-dimensional and sparse, which challenges the in-house enterprise analytics processes and models provided by multinational vendors to this major Australian bank.

We evaluate Table2Vec against the baseline models on the banking big table. The results are shown in Figure 8. Table2Vec outperforms all baselines. Table2Vec achieves the AUC and F-Score on the test data at 0.90 and 0.66, respectively, which gains 4.4% -18.9% in AUC and 18.2% 53% in F-Score over decision tree, random forest, KNN, Naive Bayes and DNN. For the two powerful methods XGBoost and GBoost, Table2Vec slightly outperforms them in terms of AUC but makes a significant improvement in terms of F-Score.

In addition, we investigate the classification accuracy of the comparison methods. Since the data is imbalanced, we evaluate the accuracy of all models in terms of the weighted accuracy. The results are shown in Figure 9, which reports that the proposed Table2Vec achieves the highest weighted accuracy. It improves 3.4% over XGBoost (the second-best method) and 24.4% over Gboost, however these two methods achieve mostly comparable AUC with Table2Vec. The results further demonstrate that Table2Vec performs significantly better than the baselines.

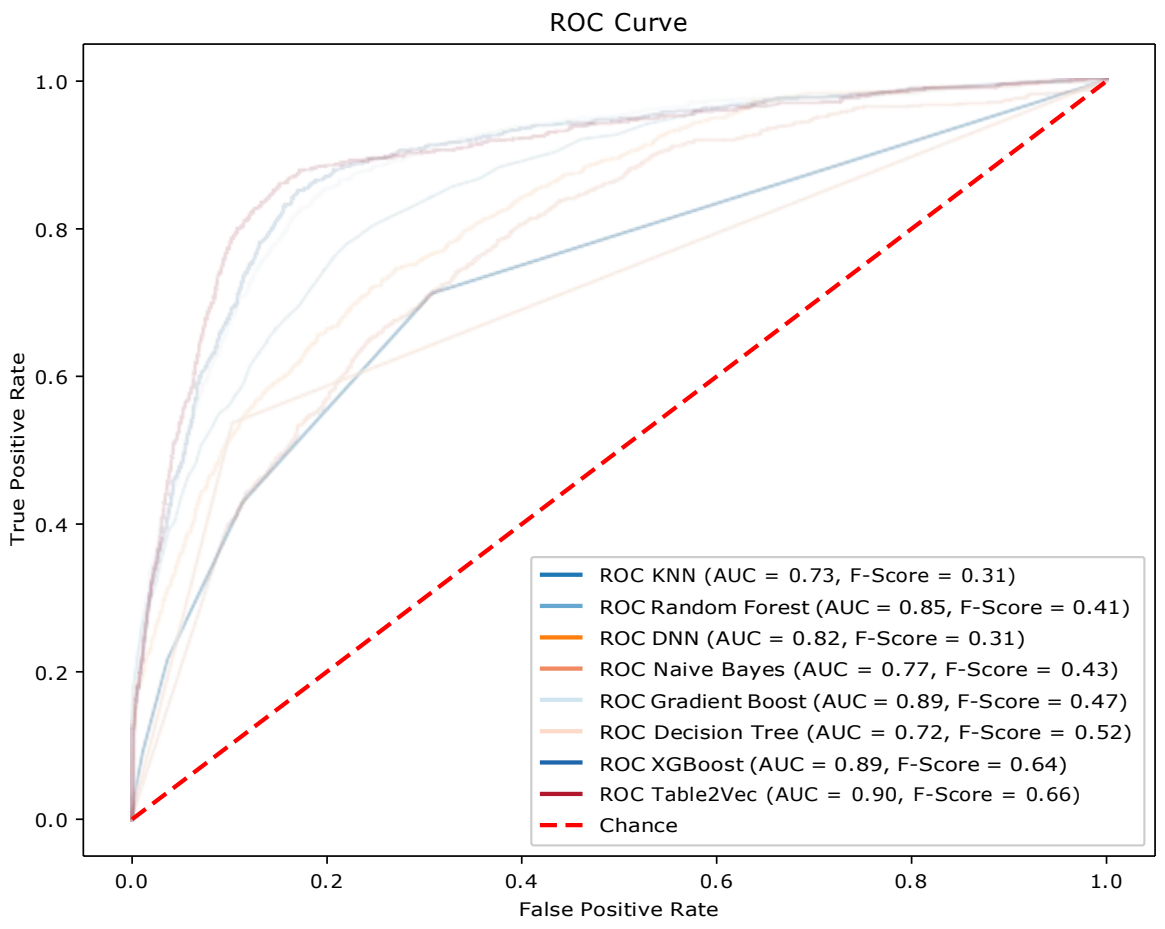

**Figure 8.** The ROC Results of Shallow and Deep Models versus Table2Vec-based Classification on the Banking Enterprise Data.

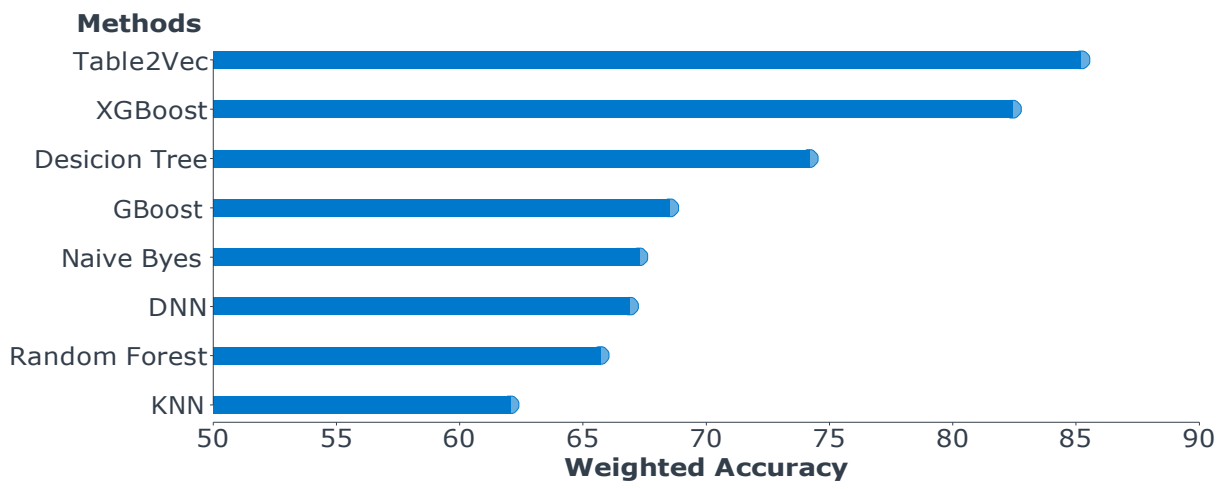

**Figure 9.** The Weighted Accuracy of Classification by Baselines versus Table2Vec-based Classification on the Banking Enterprise Data.

## Results on a Kaggle Data Set

We further test Table2Vec on a public data set[3], which has similar characteristics and complexities as the aforementioned banking data. This data set consists of 14,220,601 records of 969,182 members from an enterprise, collected from 1 January 2015 to 31 March 2017. Each member is described w.r.t. 20 features, including member meta-information, transaction information, and log information. We train the Table2Vec model on 90% of members and evaluate the Table2Vec performance on the remaining 10% of members to predict member churn.

The statistics of this data are described in Figure 10. The ratio between positive and negative labels is nearly 1:10 and the ratios of features with missing values and features with structural missingness are 52.5% and 14.5%, respectively. The ratios of static numerical features, dynamic numerical features, static categorical features, and dynamic categorical features are shown in Figure 10(d). The above statistics demonstrate that this data have the same characteristics as the banking enterprise big data, including dynamics, inconsistency, sparsity, and imbalance. We thus use this public data for the second evaluation.

The ROC results of all models on this public data are shown in Figure 11, which shows that Table2Vec significantly outperforms all competitors. Table2Vec achieves 0.87 AUC and 0.35 F-Score on the test data, respectively, which is a gain of 4.8% ∼47.5% in AUC and 9.4% 105.9% in F-Score over the other methods.

To investigate the classification performance on the data imbalance, we further report the weighted accuracy of all models in Figure 12. Table2Vec achieves impressive performance and significantly outperforms all competitors. Specifically, it improves 22.5% over the decision tree (the second best method) and 28.41% over XGBoost (with mostly comparable AUC and F-Score to Table2Vec).

To test that the representation output of Table2Vec is general and can be used for different learning tasks, we feed the Table2Vec-generated representations to the above baseline models for classification. The enhanced weighted accuracy by Table2Vec representations on the Kaggle data is shown in Figure 13. This result demonstrates the representations generated by Table2Vec can enable much better performance for both classic shallow learners and the latest deep neural models. Our experiments on the two large data sets also show that (1) both shallow and deep learners are sensitive to data characteristics; (2) it is highly challenging to learn general and robust general-purpose representations on low-quality and ultrahigh-dimensional whole-of-enterprise data; and (3) it is significant to develop universal representation learning to automatically fit specific data characteristics and produce representative, benchmarkable and shareable results for learners with different design mechanisms.

[3]https://www.kaggle.com/lclave/customer-churn

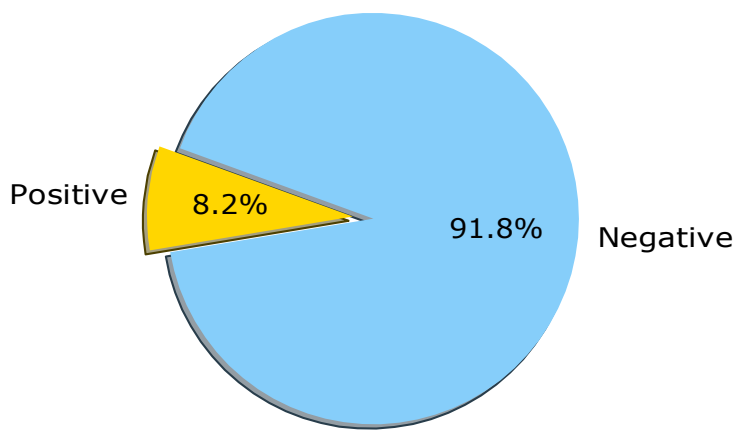

(a) Label imbalance ratio.

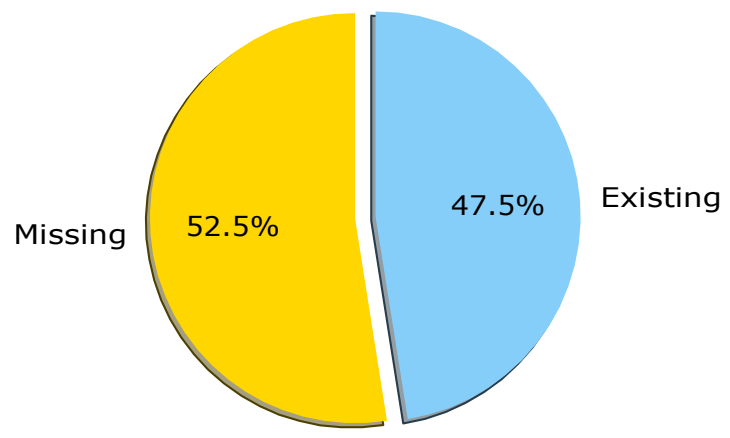

(b) Feature missing ratio.

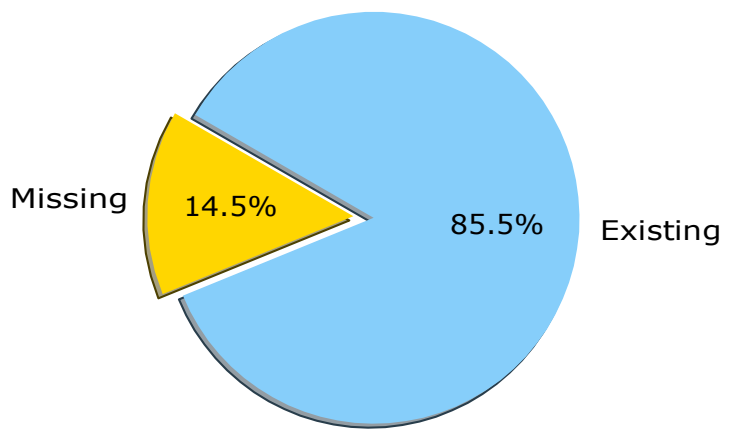

(c) Feature structural missing ratio.

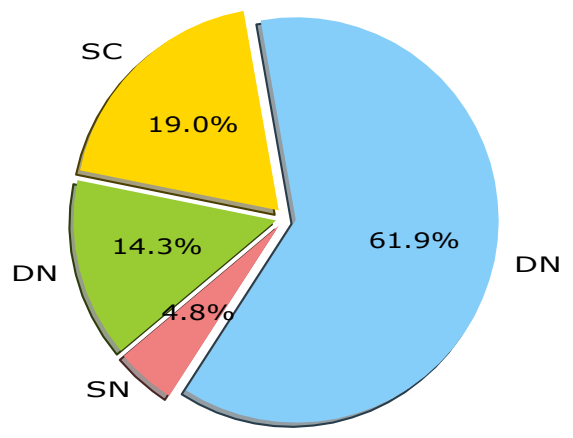

(d) Data type distribution.

**Figure 10.** Data Characteristics of a Kaggle Data Set. SC refers to static categorical data, SN refers to static numerical data, DC refers to dynamic categorical data, and DN refers to dynamic numerical data.

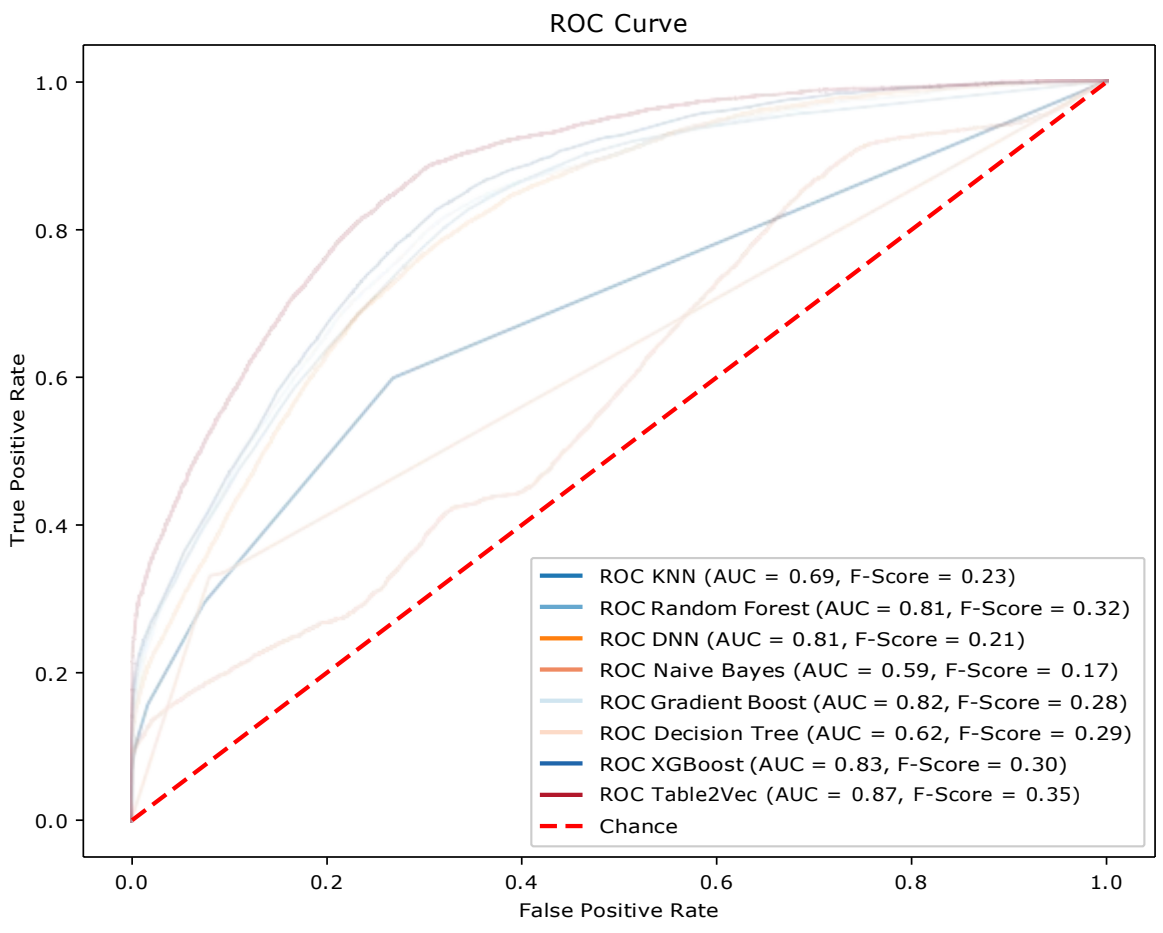

**Figure 11.** The ROC Results of Shallow and Deep Models versus Table2Vec-based Classification on the Kaggle Data Set.

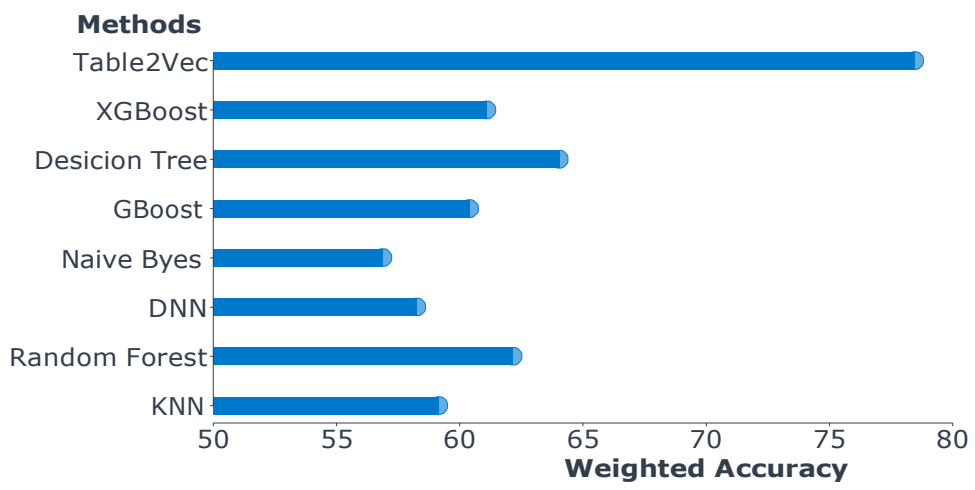

**Figure 12.** The Weighted Accuracy of Classification by Baselines against Table2Vec-based Classification on the Kaggle Data Set.

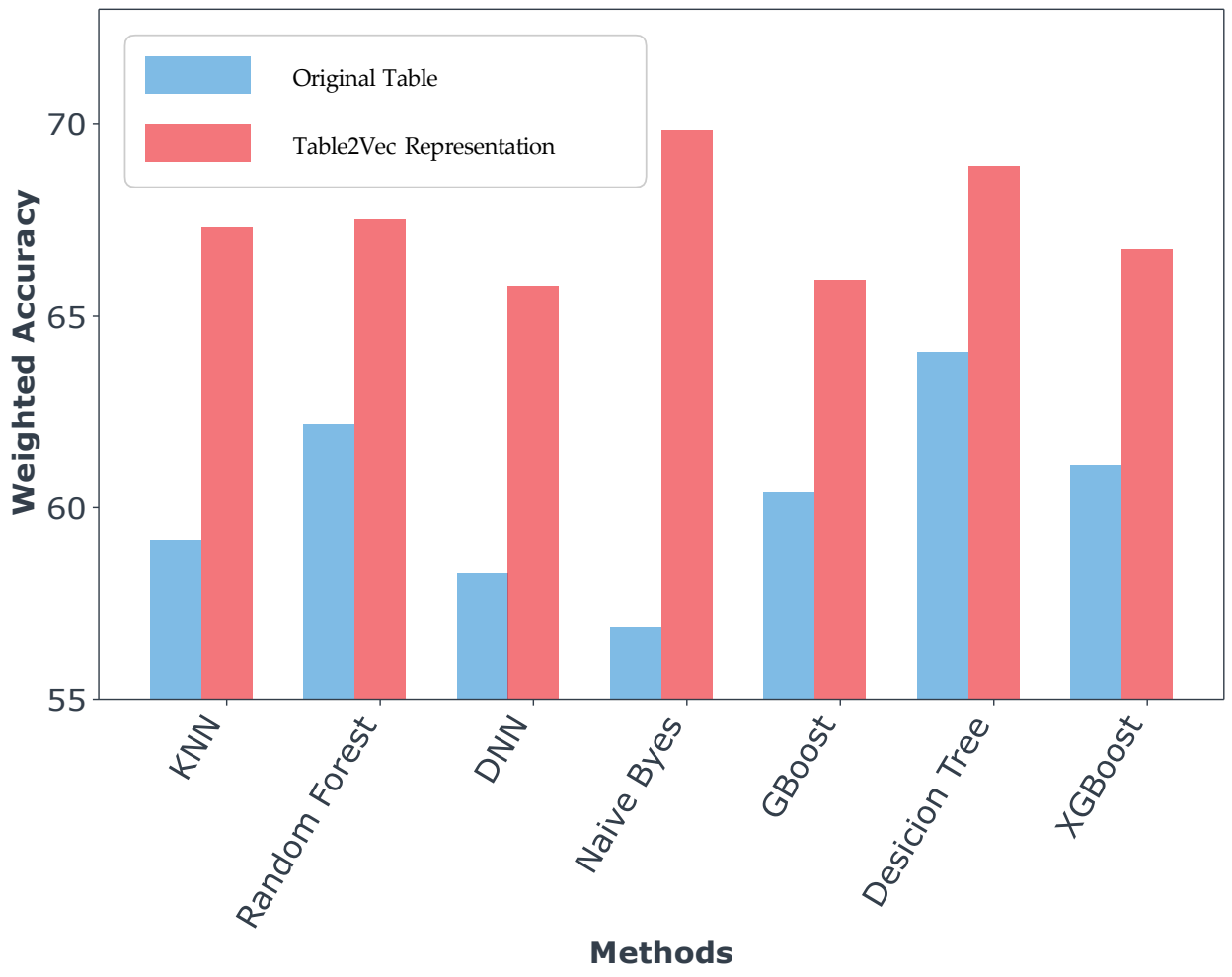

**Figure 13.** The Weighted Accuracy of Classification on the Original Table versus the Table2Vec-enabled Representation.

## Result Interpretation

We interpret the results represented by Table2Vec on the Kaggle data set to illustrate the interpretability of the Table2Vec-enabled universal representations. To better understand the reasons for member churn, we interpret the Table2Vec representations w.r.t. predicted label probability, i.e., the probability of member churn versus retention.

We first show the Table2Vec-learned customer data genomes on the Kaggle data set by the Table2Vec interpretation module, which adopts the most significant features of a customer and their intrinsic contribution customer behavior in terms of the business concern, i.e., member churn or retention, to visualize the learned customer data genomes. To illustrate the most representative data genomic factors, we report the data genomes of the most stable and unstable members found by Table2Vec in Figure 14. Here, the most stable and unstable members are those with the highest Table2Vec predicted churn and retention probability, respectively. The visualized results are consistent with our understanding of member churn that a stable member may be associated with most of the stable genomic factors but an unstable member may only have one significantly unstable genomic factor. As shown in Figure 14, the behavior sequence features, including *is_auto_renew_history*, *is_cancel_history*, and *payment_method_id_history*, of the most stable members reflect the retention patterns and make the major contribution. In contrast, the behavior sequence feature *is_cancel_history* dominates the data genomes of the most unstable member and indicates a churn characteristic. These interpretable data genomes can help the enterprise to develop personalized strategies for each customer, as they disclose the intrinsic factors, e.g., the cancellations history, related to the business concerns i.e. member churn.

We summarize the key genomes related to member retention and the churn of all members in Figure 16. As shown in Figure 16, the top-3 factors associated with member churn are *transaction cancellation history*, *payment method history*, and *automatically renew history*. For example, a significant case of member churn captured by Table2Vec was related to the continual cancellation of transactions made by the churned member. In contrast, Table2Vec predicts a member with successive transactions and frequent log information tends to be more stable. These findings are consistent with the business knowledge of customer churn, showing that Table2Vec can effectively construct the representative genomes from complex enterprise data that is general and sensitive to various business problems and learning tasks. Furthermore, the interpretation module makes the

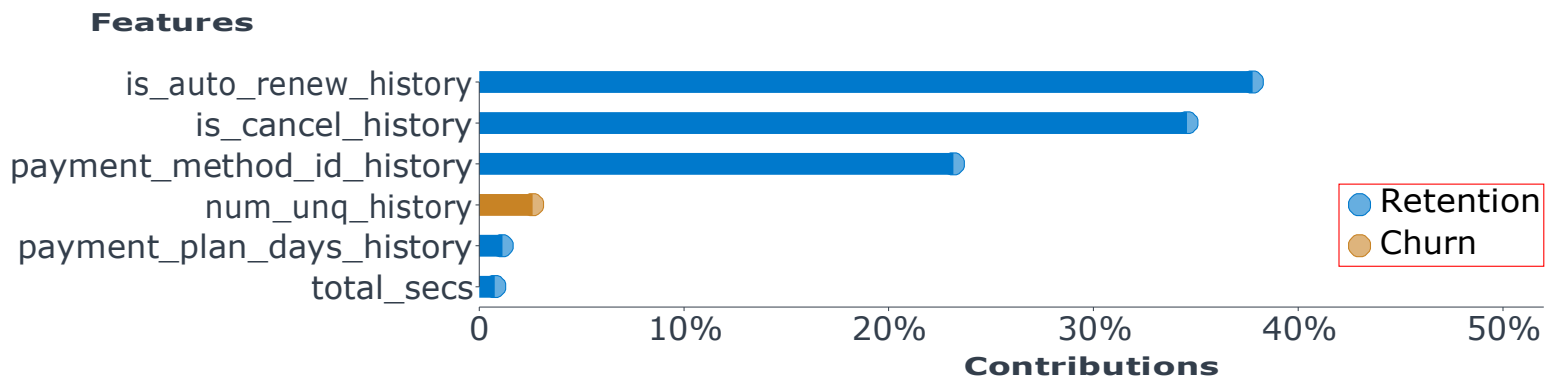

(a) The representative ‘data genomes’ of the most stable members identified from the representations learned by Table2Vec.

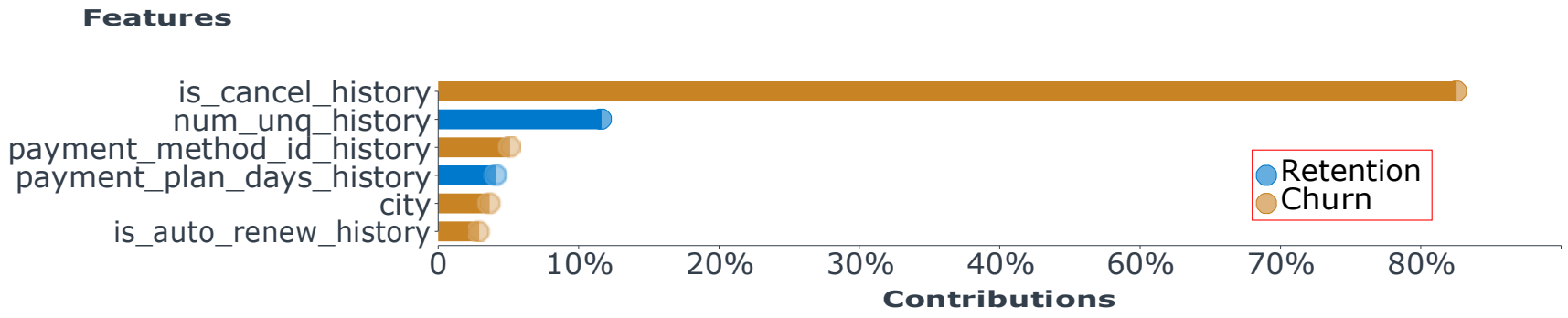

(b) The representative ‘data genomes’ of the most unstable members identified from the representations learned by Table2Vec.

**Figure 14.** The Representative ‘Data Genomes’ Discriminating Customer Churn vs. Retention in the Kaggle Data Set. The most representative genomic factors of a customer and their intrinsic contributions to customer churn or retention are discovered by the Table2Vec interpretation module.

Table2Vec representation and prediction explainable and actionable[49], contributing to explainable enterprise data science and actionable knowledge discovery for decision-making.

We further compare the Table2Vec-represented data genomes and XGBoost-selected features of customers in the public data set. Because XGBoost does not contain an interpretation module, we use the Table2Vec interpretation module to visualize their features selected by XGBoost. The results are shown in Figure 15. As revealed by Table2Vec, this customer is associated with both retention- and churn-related features, and most of the significant features are related to behavior sequences which are highly relevant to the business concern. On the contrary, all significant features selected by XGBoost are retention-related static observable features. Actually, this customer is at risk of churn, correctly predicted by Table2Vec but overlooked by XGBoost. This comparison illustrates the importance of robust representation learning to identify the deep data genomes sensitive to a business problem. In comparison with the existing methods that either ‘select’ discriminate features (such as in XGBoost) or learn ‘deep’ features (such as by DNNs), the universal representations learned by Table2Vec can be used to back-trace the representative features sensitive to its learning results, which is more representative and reliable than the hand-crafted feature engineering in shallow learners or pure black box deep learning. Our learned representations thus represent the genetic factors of customers in enterprise data.

## Discussion

Here, we briefly extend the discussion on the research issues, challenges, gaps and prospects of general-purpose enterprise data science and universal representation learning of enterprise DNA and customer data genomes. We also discuss the potential applications of universal representation learning, enterprise customer data genomes, and universal representation-based enterprise data science.

### Gaps and Opportunities

Following the discussion in the section on challenges in learning enterprise data, learning enterprise and customer data genomes for general-purpose data science involve many research challenges and pose significant research opportunities. Our proposed

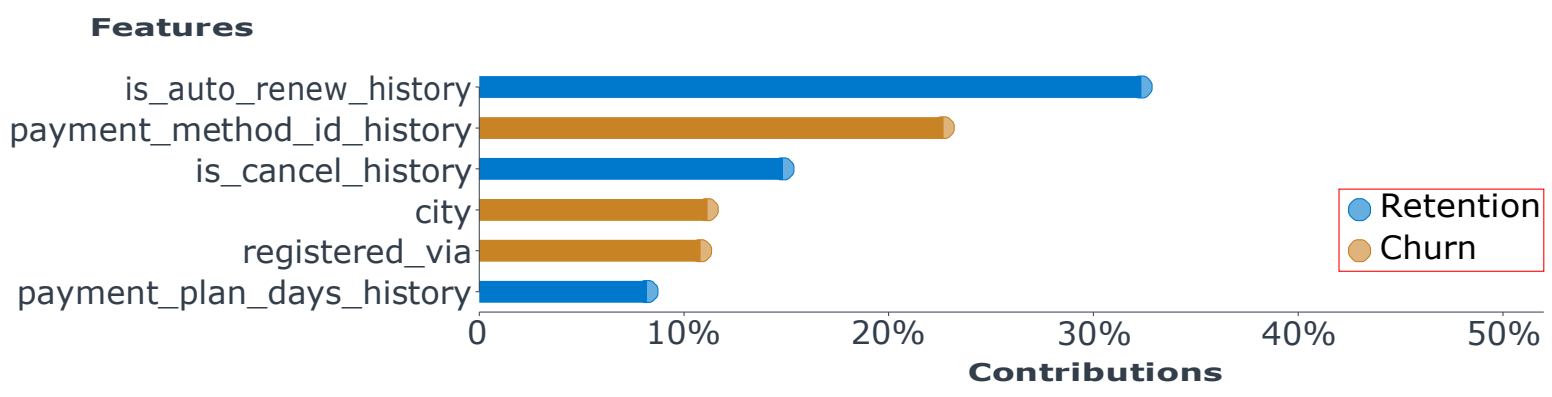

(a) The representative 'data genomes' of customers likely to churn identified from the representations learned by Table2Vec.

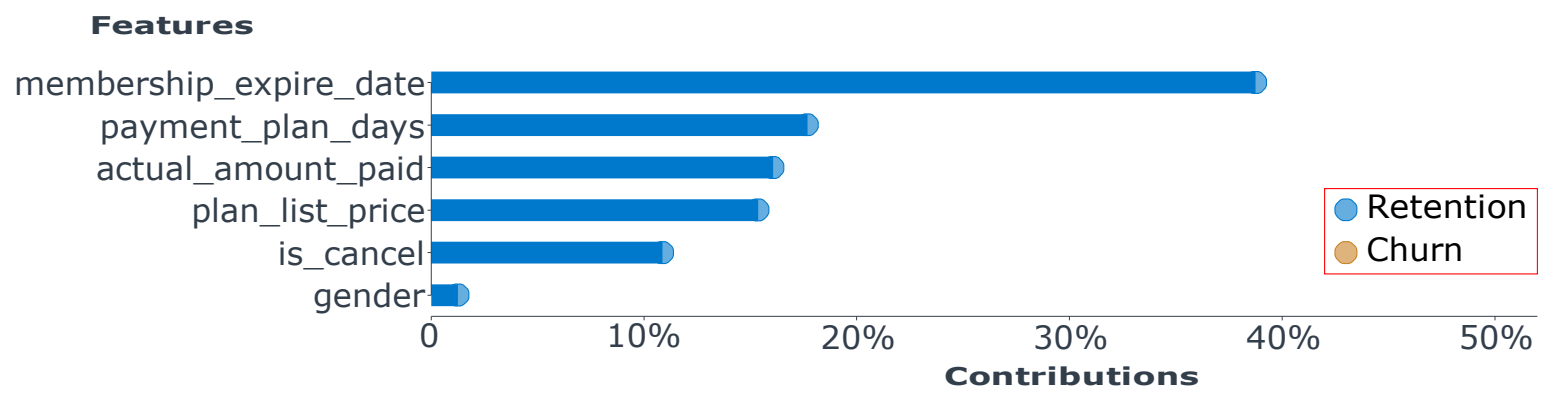

(b) The representative 'data features' of customers likely to churn learned by XGBoost.

**Figure 15.** The Comparison of Table2Vec and XGBoost in terms of Their Learned Customer Data Representations.

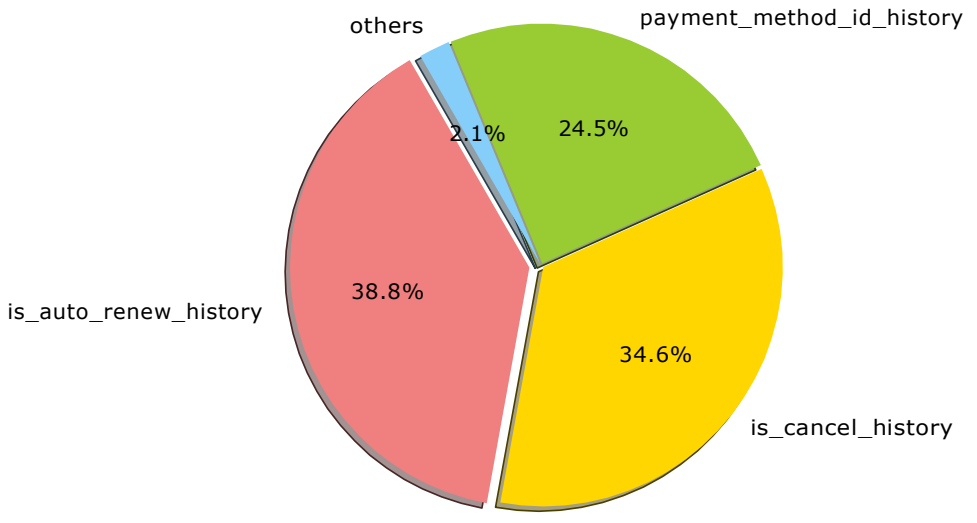

(a) The retention-sensitive representative 'genomes' identified from the representations learned by Table2Vec.

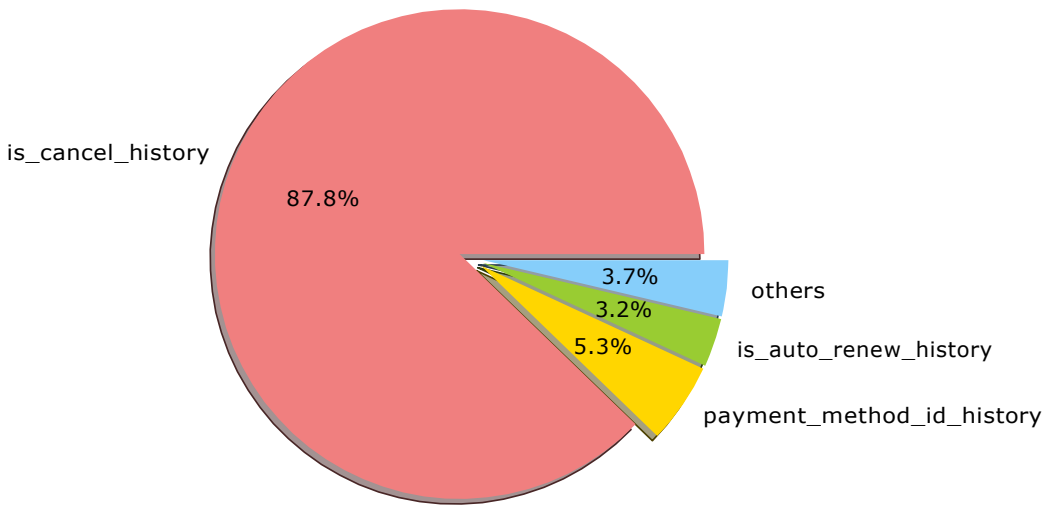

(b) The churn-sensitive representative 'genomes' identified from the representations learned by Table2Vec.

**Figure 16.** The Representative 'Data Genomes' Identified from the Representations Learned by Table2Vec.

Table2Vec only addresses some of the objectives, design thinking and expected functionalities of our proposed universal representation-based enterprise data science. It can be significantly expanded to address other issues, thinking and functionalities including but not limited to the following.

*Data DNA and Enterprise DNA*: The concept of 'data DNA' proposed in[1,10] refers to the important data-genetic information, knowledge, insights and potential of all data-based organisms such as a data-rich enterprise. Data DNA plays a role similar to the biological genomes in our human body. *Enterprise DNA* thus represents the 'genomes' of whole-of-businesses in an enterprise. *Enterprise DNA representation learning* is to extract the representative and all-round genetic information in an enterprise that can be regarded as the representations of the entire enterprise and support whole-of-enterprise data science. While enterprise data is often comprehensive and sophisticated, sufficient to build universal and representative enterprise data genomes, more research is essential to portrait the fundamental data-genetic factors in an enterprise, characterize what can be regarded as the data genomes, explore how to identify, construct, structure, formalize, learn and evaluate these data-genetic factors, and create discriminate data DNA in each specific enterprise data organism.

*Data DNA Representation Learning vs. Feature Engineering*: Feature engineering is to select, construct or mine features from a list of observable features that are sensitive to learning tasks. The representation learning of data DNA is to characterize the all-round data-genetic information hidden or observable in the data, the learned representations are built on observable features but go beyond them by also capturing other representative factors, relations, and structures, etc. The learned data genomes are not task-discriminating selected features but a mapped data expression of the important information about a physical world.

*Master Data Management vs. Universal Representation-based Enterprise Data Science*: In information systems research, master data management (MDM) aims to collect and integrate core repeatedly-used business entities (called master data) from an enterprise's business processes and systems as a high-level and reliable common data for managing data structures, system architectures, data governance, processes and quality[36]. MDM shares different concepts, objectives and tasks from our proposed universal representation-based enterprise data science. Universal representation-based enterprise data science aims to learn universal, representative, reliable and shareable/benchmarkable representations on entities such as customers, and as a result, each customer has a data genome extracted from the entire enterprise data, including domain-specific transactions and master data and operational and informational metadata as well as external data if available. Specifically, Table2Vec supports automated data augmentation to automatically recognize and process enterprise data quality issues, and then focuses on representation learning to express the data genomes of target entities such as customers.

*Universal Representations of Enterprise Data*: Enterprise data consists of not only internal resources but also external repositories, including those from third parties, external data providers, relevant vendors, and chain and portfolio companies. The third-party data and external repositories could be incorporated into the learning systems through their corresponding embedding and representation learning or enterprise data integration. Universal representation learning on enterprise data has to characterize internal and external, relational and non-relational, observable and latent, static and dynamic, and homogeneous and heterogeneous data. On one hand, learning enterprise data has to represent commonly seen numerical, categorical, textual, imagery and temporal features observable in enterprise big tables and fuse them to form an all-round representation of every aspect of an object. On the other hand, universal representation learning needs to integrate representation learning and statistical learning to learn latent variables, relations and structures in enterprise data and leverage the observable features. Both observable and latent features are heterogeneous and inconsistent, learning and fusing them in a semantically and mathematically comparable representation space is challenging, which involves various specific learning issues, including heterogeneity learning, coupling learning, ultrahigh-dimensional learning, and adaptive dynamic learning. The learned universal representations should address the aims discussed in the section on enterprise DNA and form whole-of-enterprise, whole-of-business, whole-of-customer, all lifespan, and whole-of-data expressions.

*Heterogeneity Learning*: Enterprise data is embedded with serious heterogeneity. The same and different types of features may be associated with distinct value ranges, frequencies, granularities, and distributions. Heterogeneity may also be related to structural, unstructured and semantic differences, and temporal nonstationarity, etc. Learning general-purpose representations has to learn these heterogeneities, while also converting the learned representations to some similar and mathematically comparable representation space for consistent characterization and manipulation and benchmarkable learning tasks. This forms a widely-demanding area of heterogeneity learning in data science[1], particularly important and challenging when the heterogeneities are also high-dimensional and hierarchical. Since Table2Vec takes an end-to-end neural representation learning approach, for unstructured (including long to short text), sequence and imagery data, we can easily expand Table2Vec by adding recognizers and embeddings to recognize and represent them accordingly, if necessary. The learned textual, sequential and imagery representations can be concatenated to other representations, as shown in Figure 4.

*Ultrahigh-dimensional Learning*: Representation learning of enterprise data is built on thousands of observable features. Learning from ultrahigh-dimensional ($> 1,000$ dimensions) enterprise data with various data quality issues (e.g., redundant, noisy, coupled and heterogeneous features) cannot be achieved by existing techniques, which are applicable for low-dimensional and high-quality data, such as in feature engineering and representation learning. When ultrahigh-dimensional latent features are also involved (e.g., from deep neural representation), jointly learning ultrahigh-dimensional observable and latent features is a realistic challenge for enterprise data science. It needs to tackle the significant challenges in learning representations on thousands of observable features in enterprise big tables and hierarchical ultrahigh-dimensional latent variables by learning models.

*Coupling and Interaction Learning*: Thousands of features in enterprise data may be coupled and interact strongly or weakly, statically or dynamically, explicitly or implicitly, low-rankly or high-rankly, flatly (single-level) or hierarchically, and structurally or semantically with each other in terms of different coupling relationships. Such couplings and interactions may include well-quantified relations such as association, correlation, uncertainty (dependency) and causality and poor-quantified relations such as semantic relations, political and socio-economic relations, and cultural and religious connections, etc. Enterprise data science thus requires *coupling learning*[41] of these diversified interactions and coupling relationships, including but not limited to association discovery, correlation learning, dependence learning, causality learning, and latent relation learning (such as in statistical and neural models).

*Active and Dynamic Learning*: Enterprise operations and data are associated with routine (relatively stable and incremental) and emerging (including abrupt and significant) evolution and changes. Such dynamics determines that enterprise DNA also evolves over time in terms of various factors, including customer dynamics, feature dynamics, relation and interaction dynamics, structure dynamics, and the evolution of resultant consequences, impact and outcomes, etc. Dynamics may also include new customers and new products, which may incur the change of data structures and relations. Universal representation learning of enterprise dynamics raises challenges and opportunities related to adapting to these diversified dynamics. Techniques to be developed include (1) new active and dynamic learning theories and systems, going beyond individual data sources and learning tasks and managing the dynamics associated with the aforementioned heterogeneities, interactions and couplings in enterprise data; and (2) regular change detection to automatically trigger the retraining of universal representation learning.

*Multi-objective Representation Learning*: Enterprise data science is associated with diverse business goals and learning tasks, which differ from each other and evolve over time and decision demand. Universal enterprise representation learning also needs to cater for multiple objectives of business requirements, representation learning, learning tasks, and optimization[50].

*Automated Universal Representation Learning*: General-purpose representation learning has to handle various data characteristics and data quality issues. Automated universal representation learning of enterprise data requires new representation learning theories, systems and mechanisms to automatically analyze, identify, differentiate, recognize and manage diverse data characteristics and quality issues, augment data quality, and get the data ready for further downstream learning tasks.

*Ethical Enterprise Data Science*: Enterprise data science naturally involves privacy and ethics, embedded in enterprise policies, rules, operations and governance. Ethical enterprise data science has to preserve privacy and ethics while conducting data-driven discovery on enterprise businesses and data, including enabling privacy-preserving, user-friendly, accountable, interpretable and actionable learning. The accountability, interpretability (explainability) and actionability of learning, the learned representations, and the resultant findings involve respective research issues and challenges. This includes (1) supporting privacy-preserving processing by encrypting or anonymizing identifying and sensitive information; (2) providing traceable and replayable mechanisms of interpreting the resultant representations and results; (3) mapping learned representations and results to observable features, structures, relations and patterns in the data; (4) disclosing the mapping between latent variables, relations and structures and the learned representations and results; (5) interpreting and visualizing the learning results in a user-friendly and easy to explain manner in terms of business language, rules and decision-making objectives; and (6) presenting the results in the form of evidencing decision-making and supporting decision action-taking.

*Automated Augmentation of Data Quality Issues*: As discussed in the above section on enterprise data issues, enterprise data is often incorporated with various quality issues such as redundancy, noise, missing values, inconsistency, and misinformation[1]. Enterprise data science thus needs to develop automated augmentation techniques for general data quality enhancement. Automated data augmentation caters for low-quality data, with automated augmentation (1) to recognize and manage specific data characteristics and issue scenarios; (2) clean and enhance the data quality and get the data ready for further learning tasks; or (3) directly learn low-quality and dirty data by combining data quality enhancement, feature engineering and learning tasks in one loop.

*Automated Whole-of-enterprise Learning*: Enterprise data science with capabilities to address the aforementioned listed issues and opportunities can then support automated whole-of-enterprise learning. It is integrated with functions including (1) automated data quality enhancement to recognize and handle enterprise data quality issues; (2) learning universal representations from high-dimensional observable and latent data both inside and outside an enterprise; (3) processing heterogeneities in the businesses and data; (4) learning complex couplings and interactions in enterprise; (5) learning to optimize multiple objectives if required; (6) learning dynamics in businesses and data; and (7) supporting ethics and privacy-preserving enterprise learning and decision-support.

## Potential Applications

The proposed universal representation-based enterprise data science can address many gaps and issues in existing enterprise analytics. The learned universal enterprise representations and enterprise and customer data DNA can serve multifaceted purposes and applications. Below, we illustrate some of application scenarios and business use cases that may benefit from this universal presentation and enterprise DNA learning.

- Benchmarking: The general-purpose presentations learned serve as an enterprise-level benchmark of data genomes for an enterprise to (1) support various downstream learning tasks and applications; and (2) quantify, measure, compare and interpret the similarity and difference between customer profiles, preferences, values and dynamics; between service quality and customer satisfaction; between product competitiveness and performance; and between staff and their performance; etc.

[4]Interested readers may refer to https://datasciences.org/coupling-learning/ for the relevant research on coupling and interaction learning.

- Unsupervised Learning: The general-purpose presentations learned without labeling information can support the clustering of whole-of-enterprise businesses, products, services, and customers; be used to evaluate the whole-of-business products and service quality; and group all-time customer-enterprise interactions and communications. These applications can help an enterprise to build enterprise-level segmentation and categorization of their customers, products, services, marketing activities and explain the driving factors of segmentation and categorization.

- Supervised Learning: With fully and partially labelled data in enterprise big tables, the general-purpose presentations learned can enable the classification of customers and their behaviors and interactions with an enterprise; classify products and their market performance and competitiveness; and categorize services and their quality, satisfaction; etc. The classification findings may enable enterprise-level segmentation and a comparison of customers, products, services and marketing activities to explain their classification results and underlying reasons, intervene with the target classes, and predict future occurrences of target classes for specific business needs such as the churning of high-value customers or customers for targeted promotions.

- Outlier Detection: Typical issues such as cross-product and cross-domain fraud detection, risk analytics and detection and anomaly detection in an enterprise can be built on top of the learned enterprise general-purpose representations. These include detecting which high-value customers may likely lapse; what fraudulent activities and cases exist inside an enterprise, associated with various products and services, or in communications with customers; and monitoring exceptions in key risky areas in the enterprise operations; etc.

- Trend Forecasting: Forecasting models can be built on the general-purpose presentations to estimate the enterprise trending of important production indicators, customer profiles, and performance metrics. This may include trend estimation and dynamic indexing of customer number, service usage, communication level, sentiment and satisfaction, and customer value across major business lines; estimation and dynamic indexing of product demand, service usage demand, product and service quality and satisfaction in terms of different product and service categories and marketing channels; estimation of future demand increase or decrease of specific products and services; benchmarking and trend estimation of enterprise operational performance such as productivity, average revenue per staff or customer, average revenue per product or service, customer satisfaction rate, and complaint resolution rate, etc.

- Key Factor Analysis and Interpretation: The findings of the learning tasks on the general-purpose interpretations can be further interpreted by discovering the key factors and their patternable combinations (e.g., through decision tree analysis on the results) mapped to the observable data. A visualization of the results can display key latent factors and mapped observable features and their significance level in distinguishing one finding from the others.

- Actionable and Explainable Intervention: With the aforementioned key factor analysis and interpretation and the findings showing possible problems and driving factors, it is possible to generate actionable knowledge and strategies[49] to tackle the problems with effective decision-oriented actions, explainable evidence, and significant business improvement.

In fact, there are many other use cases that can benefit from extracting and learning enterprise data DNA and general-purpose customer representations. These include opportunities and examples of both strategic and tactical aspects.

## Conclusion

Enterprise data science is receiving increasing recognition in educational, industrial and governmental institutions, innovations and decision-making. Arguably, there are many intrinsic issues in the design thinking, mechanisms and functionality of existing methodologies, systems, algorithms and tools for enterprise business intelligence, enterprise analytics, and cloud-based analytics. A landscape of universal representation learning-based enterprise data science is introduced in this work to address issues in subject-oriented enterprise data science. Automated universal enterprise representation and learning produces all-round enterprise data DNA from entire businesses and data, characterizing a whole-of-enterprise, representative, benchmarkable and shareable expression and portrayal of customers in an enterprise and enabling multifaceted enterprise-wide and domain-specific downstream learning tasks. Automated and ethical whole-of-enterprise learning on low-quality enterprise data with ethics and privacy concerns emerges as a major direction for future data science, ethical AI, enterprise analytics in enterprises to transform business intelligence theories and systems and enable data-driven enterprise innovations and automated decision-making.

## Acknowledgements

We acknowledge the funding support from the Australian Research Council Discovery grant DP190101079 and Future Fellowship grant FT190100734.

## Author contributions statement

L.B.C. contributed to the design and writing, C.Z.Z. contributed to the design, experiments and writing.

## Additional information

**Competing interests** The authors declare that they have no competing financial interests.